\documentclass{article}

\usepackage[main, final]{neurips_2026}

\usepackage[utf8]{inputenc} 
\usepackage[T1]{fontenc}    
\usepackage{hyperref}       
\usepackage{url}            
\usepackage{booktabs}       
\usepackage{amsfonts}       
\usepackage{nicefrac}       
\usepackage{microtype}      
\usepackage{xcolor}         

\usepackage{wrapfig}

\usepackage{url}
\usepackage{tcolorbox}
\usepackage{colortbl}
\usepackage{enumitem}
\usepackage{wrapfig}
\usepackage{xcolor}
\usepackage{float}
\usepackage{multirow}
\usepackage{placeins}

\usepackage{graphicx}
\usepackage{subcaption}
\usepackage{algorithm}
\usepackage{algorithmic}
\usepackage{wrapstuff}

\newcommand{\mycomment}[1]{\texttt{//} #1}

\usepackage{chngcntr}
\counterwithin{subtable}{table}

\usepackage{amsmath}
\usepackage{amssymb}
\usepackage{mathtools}
\usepackage{amsthm}

\theoremstyle{plain}
\newtheorem{theorem}{Theorem}[section]

\newtheorem{lemma}[theorem]{Lemma}

\theoremstyle{definition}

\newtheorem{assumption}[theorem]{Assumption}
\theoremstyle{remark}

\title{LEAN: Library-Based Adaptation for Asynchronous, Federated Fine-Tuning}

\author{%
  Erdong Hu\thanks{Corresponding author} \\
  Department of Computer Science\\
  Rice University\\
  Houston, TX 77005 \\
  \texttt{eh51@rice.edu} \\
  \AND
  Zhimin Ding\\
  Department of Computer Science\\
  Rice University\\
  Houston, TX 77005 \\
  \texttt{zd21@rice.edu} \\
  \And
  Yuxin Tang \\
  Amazon AGI Lab\\
  San Francisco, CA\\
  \texttt{yuxintang1995@gmail.com} \\
  \And
  Chris Jermaine\\
  Department of Computer Science\\
  Rice University\\
  Houston, TX 77005 \\
  \texttt{cmj4@rice.edu} \\
}

\begin{document}

\maketitle

\begin{abstract}
    Federated fine-tuning of foundation models with low-rank adapters (LoRA) is typically reconciled by averaging locally trained adapters. In cross-device settings, where participants hold heterogeneous data and join or leave asynchronously, averaging is poorly suited: locally trained adapters are not identifiable up to permutation, and averages of adapters trained on dissimilar data can be uninformative or unstable. We propose a simple idea called federated \emph{library-based adaptation} (\emph{LEAN}), which replaces averaging altogether. The server maintains a library of rank-1 basis pairs; participants check out random subsets, fine-tune them locally as a LoRA adapter, and check them back in, so that each pair is shuffled across sites and trained on many participants’ data. Because no aggregation step is required, LEAN operates fully asynchronously and communicates only low-rank parameters. We prove that, in a synchronous setting, LEAN converges to a neighborhood of a stationary point, with the shuffle canceling data heterogeneity at first order. Across image and text fine-tuning benchmarks, LEAN outperforms averaging-based baselines in most settings with substantially less communication, and remains effective under heavy data heterogeneity, where averaging-based methods fail to reach nontrivial accuracy, and under heavy asynchrony, where they degrade substantially.
\end{abstract}

\section{Introduction}

In federated learning (FL) \citep{fedavg, fedasync, fedopt}, a set of participants collaboratively train a joint model without sharing their local data. This paper focuses on the challenging and realistic setting of a “weak federation,” a scenario characterized by a lack of coordination and significant heterogeneity. This setting, often called cross-device FL, is defined by three key constraints:

\begin{itemize}
\item System Heterogeneity: Participants, who may be mobile or edge devices, have heterogeneous hardware and network connections. They can join or leave the training process at any time. Consequently, algorithms cannot rely on global synchronization steps.

\item Communication Bottlenecks: Participants are resource-constrained, with limited memory, compute, and especially network bandwidth. This makes frequent communication, such as uploading and downloading large foundation models, infeasible and necessitates highly communication-efficient methods.

\item Statistical Heterogeneity: Each participant's local dataset is generated from a different distribution, meaning the data is non-IID (non-independent and identically distributed) across the network. This can cause significant challenges for model convergence.
\end{itemize}

A weak federation in particular captures a classic FL case where many mobile phones collectively train a model over locally stored data (e.g., images or text) \citep{fl_mobile_edge, fl_healthcare}.

We are specifically interested in parameter-efficient fine tuning (PEFT) in this setting \citep{prompt_tuning, hu2022lora, p-tuning, qlora}, due to communication bandwidth concerns in a weak federation.
The fundamental question we consider in this paper is \emph{how to reconcile local training results into a single global model during synchronization in a weak federation?}  The universal answer is typically some variant of averaging, which is used in nearly every recent proposal for PEFT in a federated environment \citep{hetlora, fedit,  freeze_A_federated_lora}.  However, as we argue in the paper, averaging is a questionable choice, especially in a weak federation.  The fundamental problem is that an average of a set of locally trained models or adapters---that can look very different in a weak federation due to the inherent heterogeneity---may not be a meaningful global model, and in fact may be quite poor. 

In this paper, we consider a fundamentally different approach that eschews averaging.  Instead, the federation maintains a set or \emph{library} of \emph{basis pairs}. Basis pairs are pairs of vectors of the form $(\mathbf{b}^{(i)}, \mathbf{a}^{(i)})$.  Assume the federation's goal is to fine-tune a matrix $\mathbf{W}$.  To take part in this task, a participant \emph{checks out} a randomly selected subset of $r$ basis pairs.  When the library consists of $L$ basis pairs $\{(\mathbf{b}^{(i)}, \mathbf{a}^{(i)}) \}_{i = 1}^L$, a participant randomly samples an index set $\mathcal{B}$ of size $r < L$ without replacement from $\{1...L\}$ and locally fine-tunes the basis pairs $\{(\mathbf{b}^{(i)}, \mathbf{a}^{(i)}) \}_{i \in \mathcal{B}}$ to serve as an adjustment for $\mathbf{W}$ via rank-$r$ LoRA training \citep{hu2022lora}.
After fine-tuning, these basis pairs are \emph{checked in}; the participant copies its updated version of the set $\{(\mathbf{b}^{(i)}, \mathbf{a}^{(i)}) \}_{i \in \mathcal{B}}$ back into the library.  The basis pairs in the library are then deployed during inference. We call this method \emph{Library-basEd AdaptatioN}, or LEAN. In LEAN, basis pairs are never updated via averaging---in fact, they are only updated locally, by each participant then \emph{randomly shuffled} to other participants for further fine-tuning. We find strong evidence this makes LEAN more robust to heterogeneity and asynchrony, compared to averaging-based approaches.

The specific contributions of our work are:

\begin{itemize}
\item LEAN is very simple and should be easy to implement in a real-life system.  

\item LEAN addresses the identifiability problem in federated PEFT (see Section 3) while still only requiring transmission of basis pairs, in contrast to FLoRA \citep{FLoRA}, which requires broadcasting a full rank matrix during synchronization.

\item We show that LEAN naturally applies to a highly asynchronous situation where the server cannot control when each participant chooses to synchronize.  Algorithmically, handling asynchrony requires little or no adjustment in LEAN, and we show experimentally that LEAN shows high accuracy in such difficult cases.

\item We show that LEAN is robust to data heterogeneity, where participants have dissimilar data.

\item Our experiments consider a wide range of datasets and settings, with tasks that include fine-tuning of foundation models for vision and text.

\item Our code is released at \url{https://github.com/huerdong/LEAN-FL}
\end{itemize}

\section{Related Work}

Fine-tuning a full model is often prohibitively expensive, especially for Large Language Models. PEFT mitigates these costs. Some common approaches include low-rank adaptation (LoRA) \citep{hu2022lora, qlora}, p-tuning \citep{p-tuning}, and diff pruning \citep{diff_pruning}, mainly in the context of language models. Such methods have also been extended to vision transformer models as well \citep{visual_prompt_tuning, vision_language_prompt_tuning, convolution_lora}. 

One of the main approaches in FL is direct averaging of trained parameters after communication, known as FedAvg \citep{fedavg}. Many variations of this approach have since been developed such as FedProx \citep{fedprox}, MOON \citep{moon}, FedAdam \citep{fedopt}, and so on, that modify local training behaviors or add additional server post-processing. 
Most existing methods for PEFT in a federated environment use some form of averaging to reconcile locally-trained models. FedIT \citep{fedit} represents direct application of FedAvg. 
SLoRA \citep{slora} takes a meta-learning approach to construct an appropriate initialization of adapter matrices in LoRA. These so-called ``primed'' LoRA weights are then used in LoRA fine-tuning with the full weights now frozen. LoRA-A2 \citep{lora_a2}, based on FFA-LoRA \citep{freeze_A_federated_lora}, freezes alternatively the $A$ or the $B$ parameter in LoRA and only averages the trained parameter. 

A commonly addressed problem in Federated Fine-Tuning literature is  \emph{client rank heterogeneity}, where the client-trained LoRA ranks are of different ranks due to differing client compute power \citep{hetlora}.  While not explicitly considered in this paper, LEAN can be adapted to this case.

We will consider FLoRA \citep{FLoRA} carefully later in the paper---FLoRA  avoids inherent computational errors incurred when naively averaging individual LoRA adapters \citep{freeze_A_federated_lora}. However, FLoRA uses averaging, and it requires broadcasting a full rank matrix to each participant. 

A number of other asynchronous FL algorithms have been proposed. A simple adaptation of FedAvg,  {FedAsync} \citep{fedasync}, computes a weight based on the staleness of a received update. 
 {FedBuff} \citep{fedbuff} maintains a $K$-sized buffer and delaying global updates until $K$ updates are received. A similar method  {FedFa} \citep{fedfa} mitigates problems with waiting delays by using a sliding window buffer.

\section{Background: LoRA, FLoRA, Fed Learning}

LoRA is a standard method for low-rank fine-tuning or adaptation.
In LoRA, we are given a model with weight matrix $\mathbf{W} \in \mathbb{R}^{b \times a}$, and the goal is to learn a low-rank $\Delta$ matrix such that the fine-tuned model uses $\mathbf{W} + \Delta$, for $\Delta = \frac{1}{r}\mathbf{B} \times \mathbf{A}^T$, with  $\mathbf{B} \in \mathbb{R}^{b \times r}$, $\mathbf{A} \in \mathbb{R}^{a \times r}$.  In our discussion, for simplicity, we will ignore any additional LoRA scaling factors.

Because rank $r$ can be small, the value of LoRA for FL is clear---it significantly decreases network transfer cost---and its use for FL has been explored  \citep{hetlora, fedit}.  The most obvious way to apply LoRA in a federation is via some variant of federated averaging  \citep{fedavg,fedprox,fednova, moon}.  
The $i$th participant trains low-rank matrix $\mathbf{B}^{(i)} \times (\mathbf{A}^{(i)})^T$ locally, and then when it is time to reconcile $n$ local models, a global $\mathbf{B}$ is computed as $\mathbf{B} = \frac{1}{n} \sum_i \mathbf{B}^{(i)}$; $\mathbf{A}$ is computed similarly.

\textbf{The challenge in federated LoRA: identifiability.} However, there is a significant problem with this approach: the fact that $\mathbf{B} \times \mathbf{A}^T$ is invariant to ``co-permuting'' the columns of $\mathbf{B}$ and $\mathbf{A}$ (If two columns in $\mathbf{B}$ are swapped, and the matching columns in $\mathbf{A}$ are swapped, the product $\mathbf{B} \times \mathbf{A}^T$ remains unchanged).  The effect of this is that there is no obvious way to force participants to learn compatible local LoRA adapters. Two participants, even with identical datasets, can learn perfect local adapters that happen to use different permutations, and so summing/averaging during reconciliation creates global  $\mathbf{B}$ and $\mathbf{A}$ matrices that are useless.  Effectively, there is an identifiability problem: for two participants $i$ and $j$ who have trained local low-rank adapters, during reconciliation it is difficult to know which columns at participant $i$ correspond with which columns at participant $j$.

In response to this, Wang et al. proposed FLoRA (\cite{FLoRA}). Here the idea is to learn each $\mathbf{B}^{(i)}$ and $\mathbf{A}^{(i)}$ locally, 
and then, during reconciliation, to compute a global adapter $\Delta = \frac{1}{nr}\sum_i \mathbf{B}^{(i)} \times (\mathbf{A}^{(i)})^T$. 
Because each local pair is multiplied \emph{before} summing or averaging, identifiability is no longer a problem.  The authors of the FLoRA paper call this ``stacking.'' 
Thus, the scheme used in FLoRA is to ship the global $\Delta$ to each participant during synchronization.  Each participant sets $\mathbf{W} \leftarrow \mathbf{W} + \Delta$.
Each participant uses this new $\mathbf{W}$ to train a new ($\mathbf{B}^{(i)}$, $\mathbf{A}^{(i)}$) pair, and the process is repeated.

\section{FLoRA Revisited}

In this section of the paper, we consider two potential drawbacks associated with FLoRA.

\subsection{Averaging Can Lead to an Unstable Algorithm}

While FLoRA's stacking mechanism addresses the identifiability problem, there is another significant question: Data heterogeneity in a weak federation. At each synchronization step, FLoRA is averaging or summing the locally-learned adapters to compute a global $\Delta$. The problem, separate from identifiability, of \emph{client drift} \citep{scaffold} is that \emph{an average of a large set of low rank models, learned over heterogeneous data, may not be useful.}  Intuitively, if the low rank models have ``all gone in different directions'' during local adaptation, the average may simply correspond to the starting point, or worse.

\begin{figure*}
  \centering
  \includegraphics[width=0.8\textwidth]{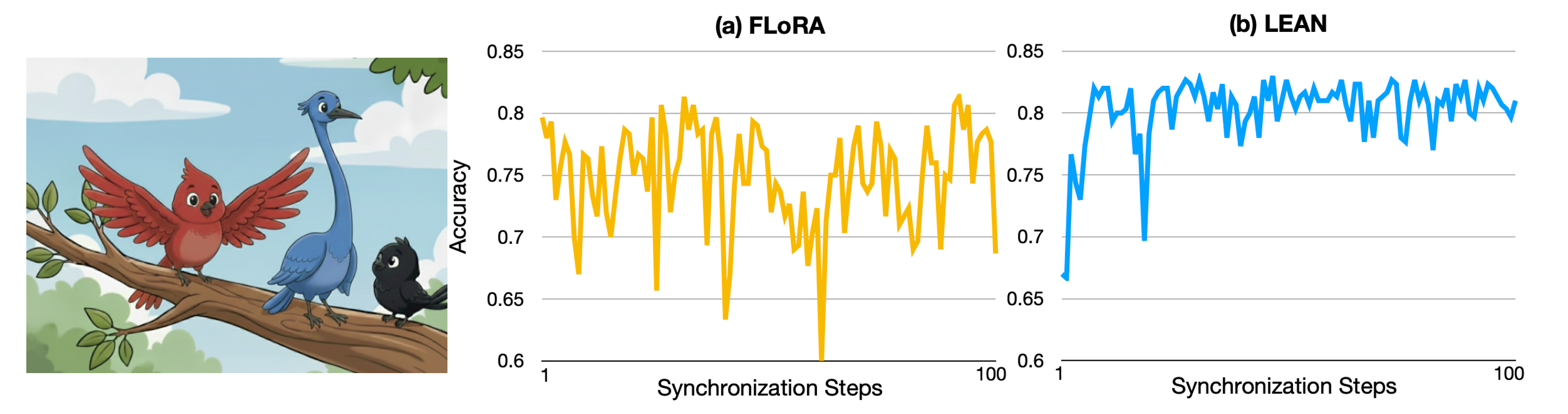}
  \vspace{-10 pt}
  \caption{Classification accuracy vs. synchronization steps for a simple synthetic task---learning to classify birds in an imaginary archipelago---with significant class imbalance across participants.}
  \label{fig:flora-vs-lean}
\end{figure*}


\textbf{Classifying birds in an imaginary archipelago.}
It is easy to construct a simple, example case that, by design, will result in questionable FLoRA performance.  Consider a synthetic dataset which consists of data describing three species of birds. A feature vector $\mathbf{x}$ has three features: bird height (in inches), and bird weight (in ounces), and bird wingspan (in inches).  Blue birds have a mean (height, weight, wingspan) of $(40, 20, 20)$, red birds have a mean of $(20, 40, 40)$, and black birds have a mean of $(5, 5, 5)$.  All feature vectors are normally distributed with standard deviation $3$.

Now, assume that those birds live on a set of islands in an archipelago. Each island has a monitoring station collecting data on 100 birds.  There are three types of islands: type 1 has 88 blue birds, 10 red, and 2 black.  Type 2 has 10 blue birds, 88 red, and 2 black birds.  Type 3 has 10 blue birds, 2 red, and 88 black.  So while it is very easy to identify what type of bird we have from the three features---the birds are all very different---the various classes are not spread evenly across the islands.

Our goal is to learn to classify birds using a simple model: $f(\mathbf{x}) = \textrm{softmax} (\mathbf{x} \times \mathbf{W})$.  $\mathbf{W}$ has three rows (as there are three features) and three columns (three bird types). Learning is to be done in a federated setting, where at each step, we send the current model to one of each of the three types of islands, and a rank-2 LoRA adapter is trained at each island.  The locally-trained adapters are reconciled using FLoRA during synchronization, and the process is repeated.

Objectively, this is an easy learning problem, or at least it would be in a centralized setting.  There is significant separation between the classes, but the class imbalance across participants is a problem for averaging, and FLoRA performs erratically. In Figure \ref{fig:flora-vs-lean}(a), we plot accuracy on a test set that consists of an equal number of each bird type, as a function of training iterations.  Accuracy fluctuates significantly over time, occasionally exceeding 80\%, but at one point falling to under 60\%.  This is concerning given the weight matrix is $3 \times 3$ and even a rank-2 adapter should be quite powerful.  

\subsection{High Communication Cost}

Another practical concern is associated with FLoRA's implementation: communication costs.
Each FLoRA participant must receive a new $\Delta$ at each training round.  The naive approach is to simply send all of the ($\mathbf{B}^{(i)}$, $\mathbf{A}^{(i)}$) pairs to each participant.  This is fine if the total size of the pairs is small, but for realistic scenarios it will not be.  A standard LoRA adapter rank is 64 or 128, and in a small foundation model that one might find in cross-device ML, the size of a weight matrix $\textbf{W}$ to be adapted is likely not much larger than $1024 \times 1024$. This means that if there are more than 8 participants, it is as costly to send all ($\mathbf{B}^{(i)}$, $\mathbf{A}^{(i)}$) pairs as it is to simply send the full update matrix $\Delta$, as $64 \times 1024 \times 2 \times 8 = 1024 \times 1024$.  Here, $64$ is the rank of the adapter, $64 \times 1024$ is the size of the adapter, $\times 2$ is because we need to send the pair, and $8$ is the number of participants.

In realistic scenarios, \emph{it is best to simply send $\Delta$ to each participant directly} (or else the current $\mathbf{W}$ matrix could be maintained at the server, and that could be sent to each participant instead).  In other words, FLoRA is performing a full-rank update, and cannot leverage low rank to save communication cost compared to full rank federated training, at least when the updated model is broadcast.

\section{Library-Based Adaptation}

\subsection{A Simple Library-Based Approach}
If indeed the cause of this behavior is FLoRA's reliance on averaging, it is reasonable to ask: How can we avoid averaging in federated adaptation?  One idea is to avoid exchanging a global average, and instead simply exchange low-rank matrices directly.  If participant $i$ computes the low-rank pair $(\mathbf{B}^{(i)}, \mathbf{A}^{(i)})$, at a synchronization step it can simply send this pair to another participant, perhaps randomly chosen, for subsequent training. The intuitive argument as to why this might work is that if a participant that has predominantly red birds sends its low-rank matrices directly to a participant with mostly black birds (for example), the receiving participant will further fine-tune the low-rank matrix with the new data, tailoring the model to the presence of the black birds. Because different individual pairs visiting multiple sites across training rounds, so-called \emph{shuffling}, each sees data from many participants.  In our example where we compute three low-rank updates at each synchronization round, we would end up circulating three updates throughout training; when it is time to publish the model, we could simply select one of the low-rank updates, or perhaps average all three at the very end.

\subsection{Maintaining a Library of Basis Pairs}\label{sec:lean_alg}

One obvious problem with this idea is that for any low-rank pair $(\mathbf{B}^{(i)}, \mathbf{A}^{(i)})$ to collect information from $n$ different participants, it will have to circulate through at least $n$ different synchronization steps. This means that many synchronization rounds may be necessary to obtain a reasonable model.

We can alleviate this problem by not circulating low-rank matrices, but by instead circulating \emph{basis pairs}. The federation maintains a set, or a \emph{library} of pairs of vectors of the form $(\mathbf{b}^{(i)}, \mathbf{a}^{(i)}), \mathbf{b}^{(i)}\in \mathbb{R}^{b}, \mathbf{a}^{(i)}\in\mathbb{R}^{a}$ which we designate the \emph{basis pairs}.  Assume the federation's goal is to fine-tune a matrix $\mathbf{W}$.  During a synchronization step, a participant \emph{checks out} a randomly selected subset of basis pairs. The library consists of $L$ basis pairs $\{(\mathbf{b}^{(i)}, \mathbf{a}^{(i)}) \}_{i = 1}^L$, and a participant randomly samples an index set $\mathcal{B}$ of size $r < L$ without replacement from $\{1,...,L\}$.  
The participant then locally fine-tunes the basis pairs using weight matrix $\mathbf{W} + \Delta$ with $\mathbf{W}$ fixed and 
$\Delta = \frac{1}{r}\sum_{i  \in \mathcal{B}} \mathbf{b}^{(i)} \times (\mathbf{a}^{(i)})^T$.  
After fine-tuning, at the next synchronization step these basis pairs are \emph{checked in}; the participant copies its updated version of the set $\{(\mathbf{b}^{(i)}, \mathbf{a}^{(i)}) \}_{i \in \mathcal{B}}$ back into the library.

Note that each basis pair that is checked out was likely last trained on a different participant.
Thus we might expect much faster adaptation/mixing when circulating basis pairs as opposed to full low-rank matrices $(\mathbf{B}^{(i)}, \mathbf{A}^{(i)})$. When the time comes to publish a model, we can randomly select $n$ pairs from the library to compute the global $\Delta$, or else use $\Delta = \sum_{i=1}^L \frac{1}{L} \mathbf{b}^{(i)} \times (\mathbf{a}^{(i)})^T$ (we use the latter in the experiments in this paper).

We call this method \emph{Library-basEd AdaptatioN}, or LEAN.
To show the potential advantage of the LEAN approach, we run the same experiment with blue, red, and black birds, using $r = 2$ for each participant and $L = 6$ basis pairs in the library.  As we see in Figure \ref{fig:flora-vs-lean}(b), LEAN is generally more accurate and more stable, even though (a) both methods train rank 2 local adapters, and (b) LEAN has less communication cost, as it never needs to communicate a full rank matrix to a participant.  

The uniform random shuffling of basis pairs across multiple sites trains each of the $L$-pairs over all sites in expectation. Since the pairs are maintained rather than aggregated, unlike FLoRA, each individual rank-$1$ pair over many rounds can ``see" the global objective, mitigating the effects of data heterogeneity. Notably, LEAN's advantage \emph{is a direct consequence of the skew in class distribution across participants.} Were there no skew, averaging would be far more meaningful and LEAN's advantage may vanish. In Appendix \ref{sec:convergence}, we provide theoretical justification for this observation as well as the shuffling mechanism, which cancels heterogeneity at first order in expectation.  

\begin{algorithm}[t]
\caption{LEAN: Federated Learning with a Library} \label{alg-LEAN}
\label{alg:fed_basis}
\footnotesize
\begin{algorithmic}[1]
\STATE \textbf{Server-side data:} Library size $L$, global basis pair library $\mathbf{B} \in \mathbb{R}^{b \times L}$, $\mathbf{A} \in \mathbb{R}^{a \times L}$, checkout status vector $\mathbf{c}$ of length $L$ initialized to zeros; standard checkout size $r$
\STATE \textbf{Shared data:} Global model weights $\mathbf{W}$

\vspace{0.1cm}

\FUNCTION{checkOut()} 
    \STATE \mycomment{Server-side execution, called by a participant.}
    \STATE Let $\mathcal{I} \leftarrow \{i | \mathbf{c}_i = 0\}$ 
    \STATE \mycomment{Compute the set of unused bases}
    \IF{$|\mathcal{I}| < r$ } 
        \STATE \mycomment{Check if there are enough unused bases}
        \STATE $L' \leftarrow L + r - |\mathcal{I}|$ \mycomment{Not enough, so extend the set}
        \STATE Append $(L' - L)$ columns to $\mathbf{B}$; init by copying random existing columns from $\mathbf{B}$
        \STATE Append $(L' - L)$ columns to $\mathbf{A}$; init by copying random existing columns from $\mathbf{A}$
        \STATE Append $(L' - L)$ scalar values to $\mathbf{c}$; init to $0$
        \STATE Add $\{L+1, ..., L'\}$ to $\mathcal{I}$; $L \leftarrow L'$ 
    \ENDIF
    \STATE Sample an index set $\mathcal{B}$ of size $r$ without replacement from $\mathcal{I}$
    \STATE Set $\mathbf{c}_i \leftarrow 1$ for all $i \in \mathcal{B}$ 
    \COMMENT{Remember these are checked out}
    \STATE Let $\mathcal{P} = \{(\mathbf{B}_i, \mathbf{A}_i) \}_{i \in \mathcal{B}}$ \COMMENT{We will return the sampled columns from $\mathbf{B}$ and $\mathbf{A}$}
    \STATE \textbf{return} $(\mathcal{P}, \mathcal{B})$ 
    \COMMENT{Return basis pairs and their original indices}
\ENDFUNCTION

\vspace{0.1cm}

\FUNCTION{checkIn($\mathcal{P}, \mathcal{B}$)} 
    \STATE \mycomment{Server-side execution, called by a participant.}
    \STATE \textbf{Input:} Updated basis pairs $\mathcal{P}$ and their indices $\mathcal{B}$.
    \FORALL{index $i \in \mathcal{B}$} 
        \STATE \mycomment{Loop through all pairs to check in}
        \STATE Remove a pair $(\mathbf{b}, \mathbf{a})$ from $\mathcal{P}$
        \STATE $\mathbf{B}_i \leftarrow \mathbf{b}$; $\mathbf{A}_i \leftarrow \mathbf{a}$; $\mathbf{c}_i \leftarrow 0$ \mycomment{Insert the pair back into the library}
    \ENDFOR
\ENDFUNCTION
\vspace{0.2cm}
\FUNCTION{enterFederation($\mathcal{D}$, $r$, $\lambda$)} 
    \STATE \mycomment{Participant-side execution.}
    \STATE \textbf{Input:} Participant's local dataset $\mathcal{D}$, desired basis set size $r$, proximal weight $\lambda$
    \STATE // \textit{Step 1: Check out basis pairs from the server.}
    \STATE $(\mathcal{P}, \mathcal{B}) \leftarrow \text{server.\textsc{checkOut}}()$.
    
    \STATE // \textit{Step 2: Locally fine-tune the checked-out basis pairs (LoRA).}
    \STATE Construct a loss function $\mathcal{L}$ over local data, parameterized on $\mathbf{W} + \frac{1}{r}\sum_{(\mathbf{b}, \mathbf{a}) \in \mathcal{P}} \mathbf{b} \times \mathbf{a}^T$
    \STATE $\mathcal{P}_0 \leftarrow \mathcal{P}$  \mycomment{$\mathcal{P}_0$ is initial parameter state used for proximal penalty}
    \STATE $\mathcal{P}' \leftarrow \text{Train}(\mathcal{L}, \mathcal{P}, \mathcal{P}_0, \mathcal{D}, \lambda)$ \mycomment{$\mathbf{W}$ is fixed; proximal penalty $\frac{\lambda}{2} ||\mathcal{P} - \mathcal{P}_0||^2_2$ used during training}
    
    \STATE // \textit{Step 3: Check in the updated basis pairs to the server.}
    \STATE $\text{server.\textsc{checkIn}}(\mathcal{P}', \mathcal{B})$
\ENDFUNCTION

\FUNCTION{getWeightMatrix()} 
\STATE \mycomment{Server-side execution; access to current weight matrix.}
\STATE \textbf{return} $\mathbf{W} + \frac{1}{L} \sum_{i = 1}^L \mathbf{B}_i \times \mathbf{A}_i^T$
\ENDFUNCTION
\end{algorithmic} 
\end{algorithm}

\subsection{Full Deployed LEAN Algorithm}

\begin{figure*}[t]
  \centering
    \includegraphics[width=0.8\textwidth]{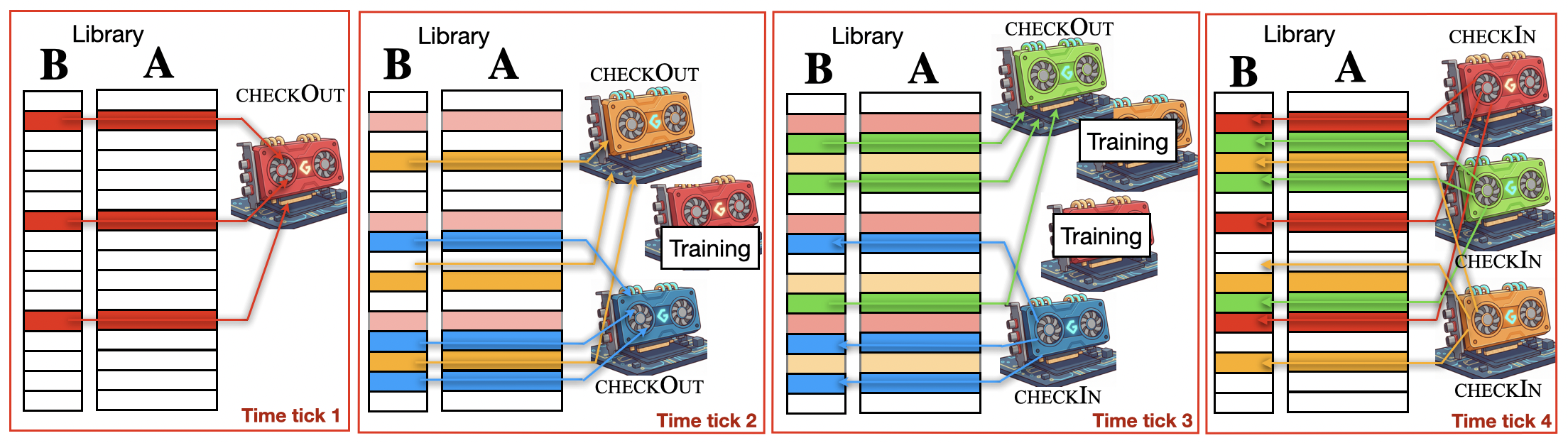}
  \caption{Asynchronous progression of the LEAN algorithm.  At time tick 1, a single participant (red) checks out $r = 3$ of $L = 16$ basis pairs.  At time tick 2, the red participant is training, and two additional sets of basis pairs are checked out.  At time tick 3, the blue participant checks in its pairs, another participant (green) enters, and two participants (red and orange) are still training. At time tick 4, all participants check in the remaining basis pairs. }
  \label{fig:asynch}
\end{figure*}

The full LEAN algorithm is given as Algorithm \ref{alg-LEAN}.  Note that in the general case, LEAN can work in a fully asynchronous setting, due to the fact that no averaging is required. When a participant is done with local fine tuning, it simply calls \textsc{checkIn} on the server to return the fine-tuned basis pairs, and they are copied into the library.  When it is ready for more work, a participant calls \textsc{checkOut} on the server to obtain a different set of basis pairs to fine-tune.  At any time, \textsc{getWeightMatrix} can be called to obtain the current, fine-tuned matrix. Asynchronous execution is demonstrated in Figure \ref{fig:asynch}.

A few points regarding Algorithm \ref{alg-LEAN} bear a bit more discussion.  First, note that the library is initialized to size $L$, but its size may be increased dynamically during \textsc{checkOut} if there are not enough basis pairs to satisfy client demand.  Second, note that during training, a proximal term may be used (if $\lambda > 0$; see line 37) to limit the amount to which a given participant can modify a basis pair.  We have found that this may be useful, especially later during training, and when data across participants are very heterogeneous.  When a basis pair is shipped to a ``unique'' participant, that participant can pull it away from the globally optimal value, and a proximal term can help.

During training, the library may grow quite large, and 
when deploying LEAN-trained models for inference, naively distributing the full basis library for inference is infeasible. As in standard implementations of LoRA \citep{hu2022lora}, to facilitate inference, it is best to distribute/deploy the adapted weight matrix itself (this updated matrix is returned directly by getWeightMatrix() in Algorithm 1). With this implementation, the cost of communicating or deploying the learned model is no different from the base model, no matter the federation/library size.

\section{Experiments}

In our experiments, we test the ability of LEAN to fine-tune modern transformers in a federated environment.  As our focus is on learning in a weak federation, we will specifically consider experiments examining the effects of data/statistical heterogeneity, asynchrony, and limited communication bandwidth.

%

\textbf{Tasks considered.}
We consider two standard fine-grained image classification datasets/tasks: Oxford VGG 102-Flowers (Flowers) \citep{flowers} and CUB-200-2011 Birds (Birds) \citep{birds}.  We also consider CIFAR100 \citep{cifar}.
Finally we consider the ChemLLMBench \citep{chembench} Reaction Prediction task (ChemRxnPred) for text-based processing.  
In the ChemRxnPred task, we are given chemical reactants in SMILES format\citep{smiles} and must generate the predicted SMILES format of the reaction outputs.

\textbf{Transformers tested.} For image classification, we use a ViT-Huge \citep{vision_transformer} model pretrained on ImageNet-21k \citep{imagenet}. The full-rank transformer blocks have rank $R=1280$ while our local training LoRA uses $r= 128$. There is an additional classification head of the same rank. For language tasks we use Qwen-3 0.6B \citep{qwen3} which has $R=1024$ blocks. Here we use $r=64$ local training rank for LoRA. 

In a federated setting, it may not be realistic to tune hyper-parameters to a specific FL task. Hence, we do not tune hyperparameters for each task.  We choose reasonable parameters and use those parameters for all tasks using the model.
Additional details of model hyperparameters are provided in the Appendix. 

\textbf{Data/statistical heterogeneity.}  Dirichlet sampling \citep{noniid_silos, fedopt} is a standard method of non-i.i.d. dataset generation. In the image tasks, we use a variant of Dirichlet sampling to partition images across participants, according to their class labels.  We use a Dirichlet parameter $\rho=1$ for ``lightly'' skewed data (that is, some classes are more prevalent on each participant, but there is not significant skew) and $\rho=0.01$ to produce heavy skew, where participants have data mostly from one or two classes.  In the language task (with no class labels) we partition data uniformly at random.




\textbf{Simulating asynchrony.}
Some of our experiments assume stochastic departures and arrivals.  Assume there are $N$ participants. At each time tick, a non-participating  site becomes a participant with probability $p=\frac{1}{N}$, and performs a \textsc{checkOut}. It then generates a duration to \textsc{checkIn} sampled from a Pareto distribution, which is governed by the scale factor $m$ and shape factor $\alpha$.  We use a scale factor $m=25$. $\alpha=1.16$ produces the ``classic'' 80-20 Pareto curve which we use as a ``light'' level of asynchrony. $\alpha=0.9$ flattens the curve which results in a difficult case where stragglers are more prevalent; and we designate this the ``heavily asynchronous'' setting.

\textbf{Federated learning methods tested.}
\textit{FedAvg with LoRA}: An approach that averages the local LoRA parameters learned \citep{freeze_A_federated_lora}. 
\textit{FedProx with LoRA}: The same, but using a proximal penalty term \citep{fedprox}. \textit{FLoRA}: the stacking-based approach discussed in Section 3 \citep{FLoRA}.  \textit{LEAN}: as described in this paper. \textit{LEAN w/o Prox}: LEAN with a 0 proximal weight. \textit{FLoRAProx}: FLoRA with a proximal term. We consider an additional baseline \textit{SLoRA} \citep{slora} in the Appendix of the paper. 

In order to adapt non-LEAN methods to the asynchronous setting, we use FedFa \citep{fedfa} where updates are averaged using a sliding window. In experiments with $N=10,25$ participants, we choose window length $5$, while for $N=100$ we use $20$.  

As a baseline, we will also consider centralized, non-federated fine tuning (``Central'').

\begin{table}[h]
   \begin{wrapstuff}[l]
        \begin{minipage}{0.45\textwidth}
        \centering
        \footnotesize
      \vspace{-10pt}
        \caption{Comparison of methods in a synchronous setting training on the VGGFlowers dataset \citep{flowers}. }
        \vspace{5 pt}
        \begin{tabular}{cccc}
        \hline
        Method & Final   & Bytes  & FLOPs \\
         &  Acc (\%) & (GB) & (Peta) \\
        \hline
            Central & 99.74\footnotemark[1]
     & N/A & N/A  \\
            FedAvg & 12.12 & DNC & DNC  \\
    
            FedProx & 17.14 & DNC & DNC \\
    
            FLoRA & 99.25 & 161.45 
             & 498.88  \\
    
            LEAN w/o Prox & 97.79 & 131.97 & 912.02  \\
    
            LEAN & 98.05 & 71.11 & 491.09  \\ 
            \hline
        \end{tabular}
        \label{tab:baseline_compare}
       \end{minipage}
    \end{wrapstuff}
\end{table}

\vspace{-20 pt}

\textbf{Statistics reported.}
In experiments where a single ``accuracy'' number is reported, it is computed as the greatest average test accuracy obtained by three consecutive models during training. We also report the total bytes transferred by the federation during training and the FLOPs required to reach high accuracy. If a method reaches at least 95\% of the accuracy of the best non-Central method in the experiment, we report the bytes and FLOPs it required to reach that threshold. If it falls below the threshold but reaches at least 70\% of it, we report the bytes and FLOPs required to reach its own peak accuracy, in parentheses, and marked with a superscript BT (``below threshold''), e.g. $(155.6)^{BT}$; \emph{these values are not directly comparable to those of methods that reached the threshold.} If it reaches less than 70\% of the threshold, we mark its bytes and FLOPs DNC (``did not converge''), as it does not attain nontrivial accuracy.
In some experiments, we instead report the best mean accuracies within a nine consecutive model window, along with the standard deviation \textit{within the same window}, in order to compare the training stability. 

\footnotetext[1]{\citep{vision_transformer}}

\subsection{The Simple Synchronous Case}\label{sec:sync}




We begin with a simple case: a fully synchronous setting where all participants enter the federation at the same time, then train locally, and finish local training at the same time (in this setting, the non-LEAN methods do not need to use a queue of recent models to deal with asynchrony). We use light heterogeneity (Dirichlet parameter $\rho = 1$).

\begin{wrapfigure}{r}{0.45\textwidth}
    \begin{minipage}{0.45\textwidth}
    \vspace{-0.10cm}
  \centering
  \includegraphics[width=\textwidth]{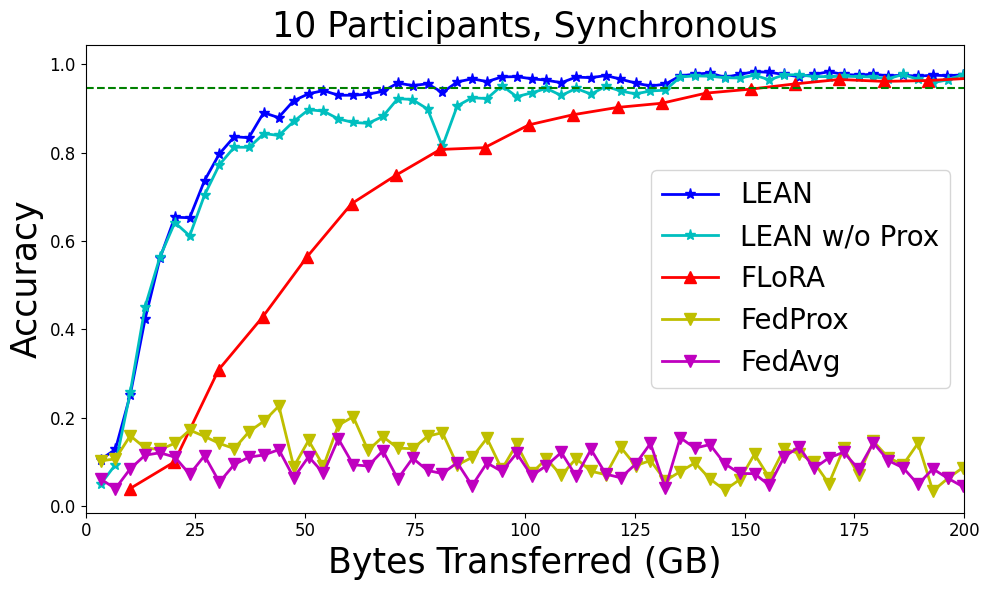}
  \caption{Accuracy as a function of bytes transferred, on 10 participants, synchronous. The dashed line indicates 95\% of centralized fine-tuning accuracy.}
  \label{fig:sync10_bytes}
  \end{minipage}
\end{wrapfigure}

Table \ref{tab:baseline_compare} gives results for the VGGFlowers data set.  In Figure \ref{fig:sync10_bytes} we show the test accuracy of the various methods on this data set as a function of the bytes transferred.

There are a few key findings. First, FedAvg and FedProx do not work well in combination with LoRA---this is not surprising, given the identifiability problem discussed earlier in the paper.  As a result, we do not consider them further in these experiments.  Also, we found that there is a significant advantage to using a proximal penalty in LEAN.  The accuracy is not that much lower than LEAN with a proximal term, but the proximal term significantly aids in convergence.  Thus we will not consider LEAN w/o Prox further, and only consider LEAN and FLoRA in subsequent experiments.  In this simple case, with little statistical heterogeneity and full synchronicity, the final accuracy of FLoRA is very good, surpassing LEAN---though FLoRA does require much more communication to converge, given the fact it must transmit full rank matrices.

\begin{table*}[h]
    \centering
    \footnotesize
    \caption{Comparison of LEAN and FLoRA for various numbers of participants with both light skewness and asynchrony. We additionally provide Central training accuracy as a reference.}
    \resizebox{\textwidth}{!}{
    \begin{tabular}{ccccccccc}
    \hline
    \multirow{2}{1em}{Part.} &  \multirow{2}{1em}{Dataset}  & Central & \multicolumn{3}{c}{LEAN} & \multicolumn{3}{c}{FLoRA}\\
    \cline{3-9}
    & & Acc. (\%) & Acc. & Bytes (GB) & FLOPs (Peta) & Acc. & Bytes & FLOPs\\ 
    \hline
     \multirow{3}{1 em}{10} &  Flowers & 99.74\footnotemark[1] & 95.18 & 44.12 & 304.4 & 89.03 & $(83.86)^{BT}$ & $(258.3)^{BT}$ \\
      &  Birds & 89.40\footnotemark[2] & 63.08 & 54.93 & 378.8 & 21.72 & DNC  & DNC  \\
      &  CIFAR100 & 94.55\footnotemark[1] & 79.74 & 47.46 & 327.8 & 1.19 & DNC & DNC \\
      \hline
    \multirow{2}{1 em}{25} &  Flowers & 99.74\footnotemark[1] & 94.77 & 71.82 & 494.2 ($1.18\times$) & 93.43 & 134.9 ($1.88\times$) & 417.3 \\
      &  Birds & 89.40\footnotemark[2] & 59.78 & 172.7 & 1192.2 & 0.78 & DNC & DNC \\
      \hline
      100 & Flowers & 99.74\footnotemark[1] &  74.73 & 288.6 & 1993.2 & 8.80 & DNC & DNC \\
      \hline
    \end{tabular}
    }
    \label{tab:lightboth}
    \vspace{-1pt}
\end{table*}
\footnotetext[2]{\citep{exploring_vit_finegrained}}

\subsection{Light Data Heterogeneity and Asynchrony}

\begin{figure}
  \centering
  \includegraphics[width=0.45\columnwidth]{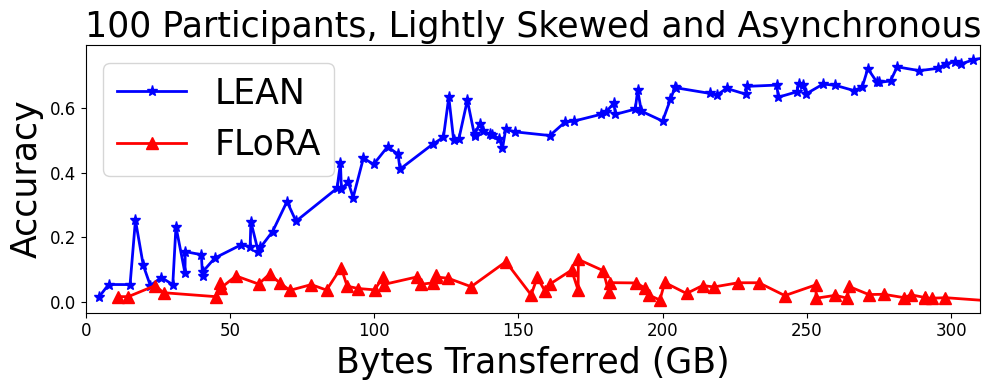}
  \includegraphics[width=0.45\columnwidth]{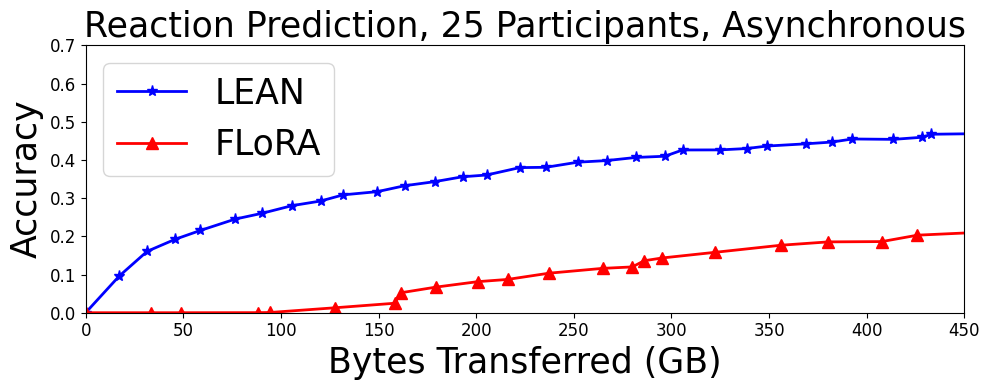}
  \caption{Comparison of LEAN with FLoRA for 100 participants on the Oxford Flowers dataset (left), and 25 participants (right) for the Reaction Prediction task (textual data). }
  \label{fig:100cli_flora_vs_lean_lightboth}
  \vspace{-0.7cm}
\end{figure}

We now consider a somewhat more difficult case: participants start and finish training lightly asynchronously.  The data are again lightly heterogeneous across participants.  Table \ref{tab:lightboth} shows results on various image classification tasks for LEAN and FLoRA in this setting.  We also plot, in Figure \ref{fig:100cli_flora_vs_lean_lightboth}, test accuracy as a function of bytes transferred for two specific lightly-asynchronous cases.

Table \ref{tab:lightboth} clearly shows that FLoRA, using a running average to handle light asynchrony, reaches the 95\% threshold only on Flowers with 25 participants, where it requires $1.88\times$ more communication than LEAN. It comes close on Flowers with 10 participants (89.03\% vs. 95.18\%), but even to reach this lower peak it uses $1.9\times$ the communication to 90.42\% threshold (95\% of LEAN's peak). In all other cases it produces poor accuracy, typically worsening with more participants. Figure \ref{fig:100cli_flora_vs_lean_lightboth} shows that FLoRA generally lags behind LEAN in terms of accuracy as a function of bytes transferred.

\setlength{\tabcolsep}{1pt} 
\begin{table}[t]
    \vspace{-20pt}
    \begin{minipage}[t]{0.45\textwidth}
    \centering
    \footnotesize
    \caption{High heterogeneity, low asynchrony.}
    \begin{tabular}{cccccccc}
    \hline
    \multirow{2}{2em}{{Part.}} & \multirow{2}{2em}{Data}   & \multicolumn{3}{c}{LEAN} & \multicolumn{3}{c}{FLoRA}\\
    \cline{3-8}
    & & Acc. & Bytes & FLOPs  & Acc. & Bytes & FLOPs\\ 
    & & (\%) & (GB) & (Peta) & (\%) & (GB) & (Peta)\\ 
    \hline
    \multirow{3}{1em}{10} & Flowers  & 69.85 & 183.8 & 1270 & 3.71 & DNC & DNC \\
    & Birds  & 25.39 & 107.8 & 743.6 & 0.62 & DNC & DNC \\
    & CIFAR & 23.35 & 224.6 & 1551.9 & 1.16 & DNC & DNC \\
     \hline
   100 &  Flowers & 23.46 & 368.0 & 2541.2 & 3.54 & DNC & DNC \\
    \hline
    \end{tabular}
    \label{tab:10cli_heavy_skew}
    \end{minipage}
    \begin{minipage}[t]{0.45\textwidth}
    \centering
    \footnotesize
    \caption{Low heterogeneity, high asynchrony.}
    \begin{tabular}{cccccccc}
    \hline
    \multirow{2}{2em}{Part.} & \multirow{2}{2em}{Data} & \multicolumn{3}{c}{LEAN} & \multicolumn{3}{c}{FLoRA}\\
    \cline{3-8}
    & & Acc. & Bytes & FLOPs  & Acc. & Bytes & FLOPs\\ 
    & & (\%) & (GB) & (Peta) & (\%) & (GB) & (Peta)\\ 
    \hline
    \multirow{2}{1em}{25} & Flowers & 94.26 & 88.09 & 608.4 & 71.30 & $(261.6)^{BT}$  & $(820.0)^{BT}$ \\
    & Birds   & 62.39 & 315.2 & 2176.8 & 0.92  & DNC  & DNC  \\
    \hline
    100 & Flowers & 78.60 & 413.8 & 2857.3 & 12.50 & DNC & DNC\\
    \hline
    \end{tabular}
    \label{tab:25cli_heavyasync}
    \end{minipage}
\end{table}

\subsection{Heavily Asynchronous and Heterogeneous}

Next, we consider the most difficult case: high levels of asynchrony or heterogeneity.  In table \ref{tab:10cli_heavy_skew}, we show some results in the case with high data heterogeneity (and low asynchrony).  In table \ref{tab:25cli_heavyasync}, we consider the case of heavily asynchronous participation (and low heterogeneity).

Table \ref{tab:10cli_heavy_skew} shows that FLoRA is particularly vulnerable to high levels of data heterogeneity across participants.  In the four cases tested, it never exceeds an accuracy of 4\%.  LEAN is also moderately affected but does much better: Flowers accuracy with 10 participants drops from about 95\% accuracy (Table \ref{tab:lightboth}) to a bit less than 70\% accuracy.  Interestingly, both methods are more robust to high asynchrony among the participants compared to levels of data heterogeneity (Table \ref{tab:25cli_heavyasync}), but LEAN is again far more robust.

\subsection{FLoRAProx}\label{sec:floraprox}

Adding a proximal term helps LEAN, might it help FLoRA?
We consider two settings: 25 participants with light heterogeneity, 10 and 100 participants with high heterogeneity.  Because the proximal term helps but FLoRAProx’s accuracy varies widely between check-ins, we report two numbers: for the nine consecutive models with the highest mean accuracy, we report the mean and the standard deviation test accuracy within that window. We run each experiment with three seeds and report the mean and standard deviation across seeds of the windowed mean accuracy and of the communication cost; the within-window standard deviation is averaged across seeds.

Table \ref{tab:floraprox} shows that with light data heterogeneity, FLoRAProx reaches a higher mean accuracy in its best window than LEAN (68.86\% vs. 62.58\%), but with more than twice LEAN’s within-window standard deviation (2.05 vs. 0.87). In a weak federation this variance matters as much as the mean: the global model may be deployed at any time, and the server cannot select a good checkpoint, since it holds no validation data. Participants who could evaluate a model with their local data may leave the federation indefinitely after a single check-in. The quality of a deployed model is therefore closer to the accuracy at an arbitrary point in training than to the best accuracy observed. We provide further ablations on this model selection issue in Appendix \ref{sec:model_selection}.

\setlength{\tabcolsep}{1pt}
\begin{table}[t]
    \centering
    \begin{minipage}{.9\textwidth}
    \centering
    \footnotesize
    \vspace{-10 pt}
    \caption{LEAN vs. FLoRAProx with low asynchrony.}
    \begin{tabular}{cccccccc}
    \hline
    \multirow{2}{2em}{Part.} & \multirow{2}{2em}{Data}   & \multicolumn{3}{c}{LEAN} & \multicolumn{3}{c}{FLoRAProx}\\
    \cline{3-8}
    & & Mean & Std & Bytes & Mean & Std & Bytes \\
    & & Acc & Dev & (GB) & Acc & Dev & (GB) \\
    \hline
    \multicolumn{8}{c}{Light Heterogeneity}\\
    \hline
    25 &  Birds & 62.58($\pm 2.09$) & 0.866 & 559.7($\pm30.79$) & 68.86($\pm3.93$) & 2.05 & 1143.8($\pm 922.3)$ \\
    \hline
    \multicolumn{8}{c}{ High Heterogeneity}\\
    \hline
    \multirow{1}{1em}{10} & Birds  & 21.54 ($\pm 1.66$) & 1.79 & 183.8 ($\pm 20.61$) & 11.91($\pm 19.51$) & 4.01 & 424.9($\pm 561.0$) \\
     \hline
    100 &  Birds & 11.68($\pm0.125$) & 1.11 & 1127($\pm200.4$) & 3.51($\pm4.04$) & 1.04 & 571.6($\pm691.4$) \\
    \hline
    \end{tabular}
    \vspace{-10 pt}
    \label{tab:floraprox}
    \end{minipage}
\end{table}

In the case of high data heterogeneity, the performance of FLoRAProx is much worse. With 10 participants, LEAN achieves a higher mean accuracy (21.54\% vs. 11.91\%) and a lower within-window standard deviation (1.79 vs. 4.01), while FLoRAProx’s standard deviation across seeds (19.51) exceeds its mean, so its outcome depends heavily on the seed. With 100 participants, FLoRAProx is unable to produce a usable model at all (3.51\%). Our conclusion is that the addition of a proximal term is able to help FLoRA to an extent, but that the effects shown in the toy example of Figure 1 still hold: the model quality tends to oscillate, and in very tough cases (as in the last row of Table \ref{tab:floraprox}), FLoRAProx is still not a viable alternative.

\section{Limitations}

Our convergence analysis covers a simplified synchronous, full-participation setting with full local gradients and no proximal term. Extending the analysis to asynchronous participation is left for future work. LEAN mitigates stragglers through library size rather than addressing them directly, and a flexible library admits adversarial behaviors not present in standard FL (Appendix D). Except for the stability comparison in Table 5, results are from single-seed runs; their corresponding conclusions mostly concern convergence versus non-convergence and large gaps in accuracy or communication, which are less sensitive to seed variation. We discuss further limitations in Appendix D.

\section{Conclusion}

In this paper, we have considered the problem of low-rank adaptation.  In particular, we have focused on the challenging problem of a \emph{weak federation}, which is characterized by system heterogeneity, communication bottlenecks, and statistical/data heterogeneity.  We have argued that the averaging-based approaches that dominate the scientific literature are perhaps of limited utility using modern models (such as transformers) in the case of a weak federation.  We have proposed LEAN, which is a new method for LoRA-like adaptation that does not use averaging. 

\begin{ack}
This research was supported by the NSF under grant numbers
2212557, 1918651, NIH CTSA award No.
UL1TR003167, and the US DOT Tier-1 UTC CYBER-CARE grant.
\end{ack}

\bibliography{lean_icml}
\bibliographystyle{plain}


\newpage

\appendix

\section{Additional Related Work}

\subsection{Non-averaging Approaches to Federated Learning}

In this work we emphasize non-averaging federated reconciliation as an alternative to averaging algorithms such as FedAvg \citep{fedavg} or FLoRA \citep{FLoRA}. A notable class of federated algorithms is decomposition approaches, such as IST \citep{ist}, FjORD \citep{fjord}, FedDropout \citep{feddrop}, or HeteroFL \citep{diaoheterofl}, where a central model is partitioned into smaller sub-networks which are distributed at client sites. These smaller models can help mitigate computation and communication costs compared to the typical FedAvg full-model broadcast as well as mitigate the impact of heterogeneous data \citep{hu2023fl_images}. However, a fundamental drawback of these partitioning methods is their lack of scalability --- because the central model is of fixed size, partitioning to large numbers of clients can fail because the sub-models can become arbitrarily or impossibly small. LEAN avoids this pitfall of previous works by taking instead a bottom-up approach where the sub-models are of fixed-size, as LoRA adapters, and the full-model --- the library --- is materialized by concatenating and reusing the sent adapters. The LoRA-based implementation of LEAN mitigates scalability concerns during inference by absorbing the library bases into the full weight matrix. We emphasize that the nonlinearity inherent in LoRA adapters requires a distinct theoretical treatment from previous works such as IST \citep{ist}.          

\subsection{Byzantine clients and Robust Aggregation}

Privacy and security are key problems in federated learning, particularly in the presence of adversarial Byzantine clients that may send detrimental model updates \citep{byzantine_bucketing, byzantine_experimental, byzantine_robust}. Various robust aggregation protocols have been proposed to mitigate Byzantine updates such as Median, TrimmedMean \citep{byzantine_robust} and Krum \citep{krum}, among many others. Notably non-averaging aggregation via order statistics is a common approach in robustly aggregated federated learning. However, a critical weakness to these approaches is their brittleness to legitimate data heterogeneity \citep{byzantine_bucketing, byzantine_experimental}. We stress that LEAN addresses problems of low-resource client connectivity and heterogeneous client data, and we assume a threat model where all participants are honest (non-Byzantine) to avoid the issue of distinguishing Byzantine clients from honest heterogeneous clients. Hence, while we acknowledge that the field of robust aggregation has also taken non-averaging approaches to federated aggregation, these methods also address a fundamentally distinct problem scope. 

\subsection{Gossip Learning}

Gossip learning \citep{hegedus} is a fully decentralized alternative to federated learning. There is no central server; instead, models are exchanged between peers, typically by having each model perform a random walk over the network. At each node, the model is updated on local data and, in most variants, merged (averaged) with the model stored there. Closely related decentralized SGD methods \citep{koloskova, lian_decentral} combine local gradient steps with averaging over a communication graph, and their analysis rests on randomized gossip and consensus \citep{chocosgd}: Repeated local averaging drives the nodes’ models to agreement at a rate governed by the mixing properties of the network.

LEAN differs from this line of work in two respects. First, LEAN uses a server-maintained library and does not assume a peer-to-peer topology. Second, and more fundamentally, LEAN performs no averaging: each basis pair is updated only by the participant that holds it, and pairs are combined additively into the adapter rather than merged into a consensus model. Nonetheless, the circulation of basis pairs bears a structural resemblance to the random-walk component of gossip learning, since each pair visits a sequence of sites and is fine-tuned at each. This suggests that tools from consensus analysis, such as the mixing time of the shuffling process or the rate at which each pair’s training history covers all sites, could be a useful lens for analyzing the library. We leave this direction to future work.

\section{Proof of Convergence} \label{sec:convergence}
Here we demonstrate the convergence of LEAN to a stationary point in the synchronous, full participation setting. 

\subsection{Notation}\label{sec:notation}
We provide the following notation:

\begin{itemize}
    \item $N$ - Number of local sites, individual sites $n\in [N]$
    \item $r$ - Local LoRA training rank/\textsc{CheckOut} size
    \item $E$ - Number of local training epochs
    \item $t$ - Synchronization round 
    \item $L\coloneq Nr$ - Full library size
    \item $\eta_t$ - Learning rate, which is a function of $t$
    \item $W\in \mathbb{R}^{b\times a}$ - Initial frozen weight
    \item $b^{(e)}_{k,t}\in \mathbb{R}^{b\times 1}, a^{(e)}_{k,t}\in \mathbb{R}^{a\times 1}$ - Library basis pairs at index $k\in [L]$ at local epoch $e\in \{0,\dots,E\}$ and on round $t$. For $e=0$, denote $b_{k,t} \coloneq b^{(0)}_{k,t}, a_{k,t} \coloneq a^{(0)}_{k,t}$, the initial values at the start of round $t$
    \item $w^{(e)}_{k,t}\coloneq b_{k,t}^{(e)}(a^{(e)}_{k,t})^T$ - LoRA update constructed from the $k$-th basis pair at round $t$ and local epoch $e$
    \item $w_{k,t}\coloneq b_{k,t}a_{k,t}^T$ - LoRA update constructed from the $k$-th basis pair at round $t$
    \item $W_t\coloneq W+\frac{1}{L}\sum_{k=1}^L w_{k,t}$ - Global model at round $t$
    \item $f_n(W_t)$ - Local objective at site $n$
    \item $f(W_t)\coloneq\frac{1}{N}\sum_{n=1}^N f_n(W_t)$ - Global objective  
\end{itemize}

Broadly, our goal is to show $f(W_t)$ can converge to a stationary point. 

Let $Sym(L)$ be the symmetry (permutation) group of order $L$. To formalize our \textsc{CheckOut} and shuffling operations, we have a permutator $\pi:\mathbb{N}\rightarrow Sym(L)$ that generates a permutation $\pi(t):[L]\rightarrow [L]\in Sym(L)$ at each round $t$. Then $\pi(t)(k)$ represents the permuted index of basis pair $k$ at round $t$. $k$ is assigned to site $n=\nu(k,t)\coloneq\lceil \frac{\pi(t)(k)}{r}\rceil$

Denote $S_{n,t}$ as the indices of basis pairs assigned to site $n$ at round $t$. Specifically, it is the $r$-partition generated by $\pi(t)$, i.e. $S_{n,t}\coloneq\{k: \nu(k,t)=n\}$. $\bigcup_{n=1}^N S_{n,t}=[L]$ and $|S_{n,t}|=r$. We define the local model $W_{n,t}$ as

\begin{align}
    W^{(e)}_{n,t} = W+\frac{1}{r}\sum_{k\in S_{n,t}} w^{(e)}_{k,t} = W+\frac{1}{r}\sum_{k\in S_{n,t}} b_{k,t}^{(e)} (a_{k,t}^{(e)})^T
\end{align}

Note that $W_{t}\coloneq\frac{1}{N}\sum_{n=1}^N W^{(0)}_{n,t}$.

\subsection{Assumptions}
We use the following set of common assumptions in FL theory \citep{fedavg_converge, fedprox, FLoRA}: 
\begin{assumption}[$L_f$-smoothness]\label{ass:l-smooth}
    $\forall n\in [N]$, $f_n$ is $L_f$-smooth, that is $\forall X, Y\in \mathbb{R}^{b\times a}$
    
    \begin{align}
        f_n(Y) \leq f_n(X) + \langle \nabla f_n(X), Y-X\rangle + \frac{L_f}{2}\|Y-X\|_F^2
    \end{align}

    And equivalently 

    \begin{align}
        \|\nabla f_n(Y) - \nabla f_n(X)\|_F \leq L_f\|Y-X\|_F
    \end{align}
    
\end{assumption}

\begin{assumption}[Bounded Gradients]
   $\forall n\in [N], X\in \mathbb{R}^{b\times a}$, there is some $G\in \mathbb{R}$
   
    \begin{align}
        \|\nabla f_n(X)\|_{F}\leq G
    \end{align}
    
\end{assumption}
We additionally introduce a bounded LoRA basis assumption, based on previous works analyzing the convergence of LoRA \citep{lora_convergence, tautological_flora_convergence}:

\begin{assumption}[Bounded basis norms]\label{ass:basis_norm}
    $\forall k\in [L], e, t$, there is some $\mathcal{C}$ that bounds the norms of the basis parameters, i.e.
    
    \begin{align}
        \|b^{(e)}_{k,t}\|^2,\|a^{(e)}_{k,t}\|^2\leq \mathcal{C}
    \end{align}
    
\end{assumption}

We make an assumption on the initialization of $b_{k,0},a_{k,0}$ to ensure correct geometric initialization:

\begin{assumption}[Clipped Gaussian Initialization]\label{ass:clipping}
    We use Gaussian sampling $b_{k}\sim \mathcal{N}(0, \frac{\mathcal{C}}{b}\mathbf{I_b}), a_{k}\sim \mathcal{N}(0, \frac{\mathcal{C}}{a}\mathbf{I_a})$, followed by a norm clipping to ensure Assumption \ref{ass:basis_norm} holds at the start as well as throughout training. 
    
    \begin{align*}
        b_{k,0} &\coloneq b_k \min(1, \frac{\sqrt{\mathcal{C}}}{\|b_k\|})\\
        a_{k,0} &\coloneq a_k \min(1, \frac{\sqrt{\mathcal{C}}}{\|a_k\|})
    \end{align*}
    
\end{assumption}

\subsection{Persistent Isometry Lemma}

Define the operator for the $k$-th basis pair as the following

\begin{align}
    \mathcal{M}_{k,t}^{(e)}(g) \coloneq ga_{k,t}^{(e)}(a_{k,t}^{(e)})^T + b_{k,t}^{(e)}(b_{k,t}^{(e)})^T g
\end{align}

We note that $\mathcal{M}_{k,t}^{(e)}$ is self-adjoint. Let $\mathcal{M}^{(e)}_{t}\coloneq\frac{1}{L}\sum_{k=1}^L \mathcal{M}_{k,t}^{(e)}$. Denote $\mathcal{M}_t\coloneq\mathcal{M}^{(0)}_{t}$ 

Our goal is to show the geometric stability of the $\mathcal{M}_t$ operator over the progression of the LEAN algorithm. 

\begin{lemma}[Persistent Restricted Isometry]\label{lemma:isometry}

For $T\in\mathbb N$ and $\delta\in(0,1)$, suppose $L\ge\frac{16}{1/b +1/ a}\log\frac{ba}{\delta}$ and $\eta_t=\eta_0/\sqrt{t+1}$ with $\eta_0\le\frac{(1/b+1/a)\,r}{32\,EG\sqrt T}$. Then $\exists \gamma >0$ such that with probability at least $1-\delta$,

    \begin{align}
        \langle g, \mathcal{M}_t(g)\rangle\geq \gamma\|g\|_F^2
    \end{align}

    for all $t\leq T, g\in\mathbb{R}^{b\times a}$
    
\end{lemma}
\begin{proof}
    Since $\mathcal{M}_{k,0}$ is a linear operator, the matrization of $\mathcal{M}_{k,0}$ is the $(ba)\times (ba)$ matrix 

    \begin{align}
        \mathbf{M}_{k,0} = (a_{k,0}a_{k,0}^T \otimes \mathbf{I}_b) + (\mathbf{I}_a \otimes b_{k,0}b_{k,0}^T)\label{eqn:matrize_init}
    \end{align}
    where $\otimes$ is the Kronecker product. We abuse notation in general and treat matrization of self-adjoint operators as interchangeable with operator itself, and the key equivalence is $\mathcal{M}_{k,t}^{(e)}(g)=\mathbf{M}_{k,t}^{(e)}g$.  

    For the base case at $t=0$, using Assumption \ref{ass:clipping}'s Gaussian initialization and then using Eqn. \ref{eqn:matrize_init}

    \begin{align}
        \mathbb{E}[\mathbf{I}_{a}\otimes b_{k,0}b_{k,0}^T] &= \frac{\mathcal{C}}{b}\mathbf{I_{ba}}\\
        \mathbb{E}[a_{k,0}a_{k,0}^T\otimes \mathbf{I}_{b}] &= \frac{\mathcal{C}}{a}\mathbf{I_{ba}}\\
        \mathbb{E}[\mathbf{M}_{k,0}] &= \mathcal{C}(\frac{1}{b}+\frac{1}{a})\mathbf{I}_{ba}\\
        \mathbb{E}[\mathbf{M}_0] &= \mathbb{E}[\frac{1}{L}\sum_{k=1}^L \mathbf{M}_{k,0}]\\
        &= \frac{1}{L}\sum_{k=1}^L\mathbb{E}[\mathbf{M}_{k,0}]\\
        \mathbb{E}[\mathbf{M}_0] &= \mathcal{C}(\frac{1}{b}+\frac{1}{a})\mathbf{I}_{ba} 
    \end{align}

    Let $\mu_{min}$ be the minimum eigenvalue of $\mathbb{E}[\mathbf{M}_0]$, which from the above is 

    \begin{align}
        \mu_{min} = \lambda_{min}(\mathbb{E}[\mathbf{M}_0]) = \mathcal{C}(\frac{1}{b}+\frac{1}{a})
    \end{align}

    Applying Assumption \ref{ass:basis_norm}, we can bound the maximum eigenvalue (Spectral norm) of $\mathbf{M}_{k,0}$ via triangle inequality

    \begin{align}
        R \coloneq\lambda_{max}(\mathbf{M}_{k,0}) &\leq \lambda_{max}(a_{k,0}a_{k,0}^T \otimes \mathbf{I}_b) + \lambda_{max}(\mathbf{I}_a \otimes b_{k,0}b_{k,0}^T)\\
        &\leq \|a_{k,0}\|^2 + \|b_{k,0}\|^2\\
        \lambda_{max}(\mathbf{M}_{k,0}) &\leq 2\mathcal{C}
    \end{align}

    Noting that $\mathbf{M}_{k,0}$ is self-adjoint, positive semi-definite, and independently initialized, we can apply Matrix Chernoff Bounds (Tropp Remark 5.3 \cite{Tropp_2011_hoeffding}) and bound the right-hand side by $\delta$. For any $\mu\leq \lambda_{min}(\mathbb{E}[\mathbf{M}_0])$,

    \begin{align}
        \mathbb{P}[\lambda_{\min}\Big(\sum_{k=1}^L\mathbf{M}_{k,0}\Big)\leq\frac{L\mu}2]&\leq ba e^{-\frac{(1/2)^2L\mu}{2R}}\leq\delta\\
        bae^{-L\mu/16\mathcal{C}} &\leq \delta\\
        L &\geq \frac{16\mathcal{C}}{\mu}\log(\frac{ba}{\delta})\label{eqn:libsize_bound}
    \end{align}

    If we let $\gamma_0\coloneq \frac{\mu_{min}}{2} = \frac{\mathcal{C}}{2}(\frac1b+\frac1a)$ and ensure that the library size L is sufficiently large from Equation \ref{eqn:libsize_bound}

    \begin{align}
        L &\geq \frac{16\mathcal{C}}{\mu_{min}}\log(\frac{ba}{\delta})\\
        L &\geq \frac{16}{1/b + 1/a}\log(\frac{ba}{\delta})
    \end{align}

    then with probability $1-\delta$

    \begin{align}
        \lambda_{\min}\Big(\sum_{k=1}^L\mathbf{M}_{k,0}\Big)&\geq\frac{L\mu_{min}}2\\
        \lambda_{\min}(\mathcal{M}_0)&\geq \frac{\mu_{min}}{2}=\gamma_0
    \end{align}
    
    
    Using Rayleigh quotient bounds,

    \begin{align}
        \langle g, \mathbf{M}_0g\rangle &\geq \lambda_{min}(\mathbf{M}_0)\|g\|^2\\
        \langle g, \mathcal{M}_0(g)\rangle &\geq \gamma_0 \|g\|^2 
    \end{align}

    as desired for the base case. 

    Now suppose for induction that for some $t$, 
    $\lambda_{\min}(M_t) \geq \gamma_0 - \sum_{s=0}^{t-1} \epsilon_s$, where

    \begin{align}
    \epsilon_s = \frac{4\mathcal{C}\eta_s EG}{r} 
    \end{align}
    
    Consider basis index $k\in S_{n,t}$ assigned to site $n$. The stepwise gradient descent update on $b_{k,t}^{(e)}, a_{k,t}^{(e)}$ is the following 

    \begin{align}
        b_{k,t}^{(e+1)} &= b_{k,t}^{(e)} - \eta_t \frac{\partial}{b_{k,t}^{(e)}}f_{n}(W_{n,t}^{(e)})\\
        &= b_{k,t}^{(e)} - \eta_t \nabla f_n(W_{n,t}^{(e)})\frac{\partial W_{n,t}^{(e)}}{\partial b_{k,t}^{(e)}}\\
        b_{k,t}^{(e+1)} &= b_{k,t}^{(e)} - \frac{\eta_t}{r}\nabla f_n(W_{n,t}^{(e)})a_{k,t}^{(e)}\label{eqn:b_local_desc}\\
        a_{k,t}^{(e+1)} &= a_{k,t}^{(e)} - \frac{\eta_t}{r}\nabla f_n(W_{n,t}^{(e)})^Tb_{k,t}^{(e)} \text{ , by similar derivation}\label{eqn:a_local_desc}
    \end{align}

    To simplify notation, we let $G_{n,t}^{(e)}\coloneq \nabla f_{n}(W_{n,t}^{(e)})$. Then the full $E$ rounds of local descent becomes 

    \begin{align}
        b_{k,t+1} &= b_{k,t} - \frac{\eta_t}{r} \sum_{e=0}^{E-1} G_{n,t}^{(e)}a_{k,t}^{(e)}\label{eqn:b_desc}\\
        a_{k,t+1} &= a_{k,t} - \frac{\eta_t}{r} \sum_{e=0}^{E-1} (G_{n,t}^{(e)})^Tb_{k,t}^{(e)}\label{eqn:a_desc}
    \end{align}

    Denote the update operator 

    \begin{align}
        \Delta\mathcal{M}_{k,t}(g) &\coloneq \mathcal{M}_{k,t+1}(g) - \mathcal{M}_{k,t}(g)\\
        \Delta\mathcal{M}_{k,t}(g) &= g[a_{k,t+1}(a_{k,t+1})^T - a_{k,t}(a_{k,t})^T] + [b_{k,t+1}(b_{k,t+1})^T - b_{k,t}(b_{k,t})^T]g\label{eqn:delta_m}
    \end{align}

    From Equation \ref{eqn:b_desc} we can get the following 

    \begin{align}
        b_{k,t}^{(e)} - b_{k,t} &= - \frac{\eta_t}{r} \sum_{i=0}^{e-1} G_{n,t}^{(i)}a_{k,t}^{(i)}\\
        \|b_{k,t}^{(e)} - b_{k,t}\|&\leq \frac{\eta_t}{r}\sum_{i=0}^{e-1} G\sqrt{\mathcal{C}}\\
        &\leq \frac{e\eta_t G\sqrt{\mathcal{C}}}{r}\label{eqn:e_residual_bound}\\
        \|b_{k,t}^{(e)} - b_{k,t}\|&\leq \frac{\eta_tEG\sqrt{\mathcal{C}}}{r}\\
        \|b_{k,t+1} - b_{k,t}\|&\leq \frac{\eta_tEG\sqrt{\mathcal{C}}}{r}
     \end{align}

    using the fact that $b_{k,t+1}=b_{k,t}^{(E)}$. Applying the identity $\| XX^T - YY^T \| \leq (\|X\| + \|Y\|)\|X - Y\|$:

    \begin{align}
        \|b_{k,t+1}b_{k,t+1}^T - b_{k,t}b_{k,t}^T \|&\leq (\|b_{k,t+1}\| + \|b_{k,t}\|)\|b_{k,t+1}-b_{k,t}\|\\
        &\leq (2\sqrt{\mathcal{C}})(\frac{\eta_t EG\sqrt{\mathcal{C}}}{r})= \frac{2\mathcal{C}\eta_t EG}{r}\label{eqn:residual_bound}
    \end{align}

    And applying a similar inequality for $a_{k,t}$ gives us the operator norm bound on $\Delta \mathcal{M}_{k,t}$ from Equation \ref{eqn:delta_m}

    \begin{align}
        \|\Delta\mathcal{M}_{k,t}(g)\|_F&\leq \|g\|_F\|a_{k,t+1}(a_{k,t+1})^T - a_{k,t}(a_{k,t})^T\|_2 \\
        &+ \|b_{k,t+1}(b_{k,t+1})^T - b_{k,t}(b_{k,t})^T\|_2\|g\|_F\nonumber\\
         \|\Delta\mathcal{M}_{k,t}(g)\|_F &\leq \frac{4\mathcal{C}\eta_tEG}{r}\|g\|_F\\
         \|\Delta \mathcal{M}_{k,t}\|_2 &= \sup_{g\neq 0}\frac{ \|\Delta\mathcal{M}_{k,t}(g)\|_F}{\|g\|_F}
        \\
        \|\Delta \mathcal{M}_{k,t}\|_2 &\leq \frac{4\mathcal{C}\eta_t EG}{r}\label{eqn:operator_drift}
    \end{align}

    The total library update operator becomes

    \begin{align}
        \mathcal{M}_{t+1} = \mathcal{M}_t + \frac{1}{L}\sum_{k=1}^L\Delta\mathcal{M}_{k,t}
    \end{align}

    Then we can apply Weyl's Inequality and Triangle Inequality

    \begin{align}
        \lambda_{\min}(\mathcal{M}_{t+1})&\geq\lambda_{\min}(\mathcal{M}_t)-\frac1L\Big\|\sum_{k=1}^L\Delta \mathcal{M}_{k,t}\Big\|_2\\
        &\geq \lambda_{\min}(\mathcal{M}_t)-\frac1L\sum_{k=1}^L\|\Delta \mathcal{M}_{k,t}\|_2\\
        \lambda_{\min}(\mathcal{M}_{t+1})&\geq\lambda_{\min}(\mathcal{M}_t)-\frac{4C\eta_tEG}{r}.
    \end{align}


    From the inductive assumption

    \begin{align}
        \lambda_{min}(\mathcal{M}_{t+1})&\geq \gamma_0 - \sum_{s=0}^{t-1}\epsilon_s - \frac{4C\eta_tEG}{r}\\
        &\geq \gamma_0 - \sum_{s=0}^{t-1}\epsilon_s - \epsilon_t\\
        \lambda_{min}(\mathcal{M}_{t+1})&\geq \gamma_0 - \sum_{s=0}^{t}\epsilon_s
    \end{align}

    which proves the inductive step. 

    Let $\eta_t = \frac{\eta_0}{\sqrt{t+1}}$ to ensure that $\eta_t\leq \eta_0$. Setting 
    
    \begin{align}
        \eta_0 \leq 
    \frac{(1/b + 1/a)r}{32EG\sqrt{T}}=\frac{\gamma_0}{2} (\frac{r}{8\mathcal{C}EG\sqrt{T}}) 
    \end{align}

    And using the identity $\sum_{s=0}^{T-1} \frac{1}{\sqrt{s+1}} \leq 2\sqrt{T}$, we derive for all $t\leq T$

    \begin{align}
        \sum_{s=0}^{T-1} \epsilon_s &= \frac{4\mathcal{C}EG}{r}\sum_{s=0}^{T-1}\eta_s\\
        &= \frac{4\mathcal{C}EG}{r}\eta_0\sum_{s=0}^{T-1}\frac{1}{\sqrt{s+1}}\\
        &\leq \frac{8\mathcal{C}EG\sqrt{T}}{r}\eta_0\\
        \sum_{s=0}^{T-1} \epsilon_s&\leq \frac{\gamma_0}{2}\\
        \sum_{s=0}^{t-1} \epsilon_s&\leq \frac{\gamma_0}{2}
    \end{align}

    Therefore

    \begin{align}
        \lambda_{min}(\mathcal{M}_t)&\geq \gamma_0 - \sum_{s=0}^{t-1}\epsilon_s\\
        &\geq \gamma_0 - \frac{\gamma_0}{2}\\
        \lambda_{min}(\mathcal{M}_t)&\geq \frac{\gamma_0}{2}
    \end{align}

    Setting $\gamma=\gamma_0/2$ satisfies the desired bound

    \begin{align}
        \langle g, \mathcal{M}_t(g)\rangle\geq \gamma\|g\|_F^2
    \end{align}

    And since the inductive step is deterministic, the probability is derived from the Matrix Chernoff's $1-\delta$. 

\end{proof}

\subsection{Proof of Descent}

We now show the main theoretical results 

\begin{theorem}\label{thm:converge}
    For any $T\geq 1$, suppose Assumptions B.1-B.4 hold, with $L=Nr$ and $L$ satisfying the hypothesis of Lemma \ref{lemma:isometry}. Let $\eta_t = \frac{\eta_0}{\sqrt{t+1}}$ where 

    \begin{align}
        \eta_0 \leq \min\{\frac{r}{G}, \frac{\gamma}{8\mathcal{C}GE^2}, \frac{r(1/b+1/a)}{32EG\sqrt{T}}\}
    \end{align}

    Over $T$ training rounds, with probability at least $1-\delta$, $\delta\in (0,1)$
    
    \begin{align}
        \frac{1}{T}\sum_{t=0}^{T-1}\mathbb{E}[\|\nabla f(W_t)\|_F^2]\leq \frac{2(f(W_0) - f^*)}{\gamma\eta_0\sqrt{T}} + \frac{16L_fE\mathcal{C}^2G(N-1)}{N\gamma} + \frac{K\eta_0(1+\ln T)}{\gamma r\sqrt{T}}
    \end{align}

where 

\begin{align}
    K\coloneq 12\sqrt3\,L_fE^2C^2G^2+2G^3EC+27L_fG^2E^2C^2+8E(E-1)CG^3
\end{align}
    
\end{theorem}
\begin{proof}
    Consider the local update on pair $w_{k,t}^{(e)}$ to $w_{k,t}^{(e+1)}$. Using Equations \ref{eqn:b_local_desc} and \ref{eqn:a_local_desc} from the proof of Lemma \ref{lemma:isometry}

    \begin{align}
        w_{k,t}^{(e+1)} &= b_{k,t}^{(e+1)}(a_{k, t}^{(e+1)})^T\\
        &= \left(b_{k,t}^{(e)} - \frac{\eta_t}{r} G_{n,t}^{(e)}a_{k,t}^{(e)}\right)\left(a_{k,t}^{(e)} - \frac{\eta_t}{r} (G_{n,t}^{(e)})^Tb_{k,t}^{(e)}\right)^T\\
        &= w_{k,t}^{(e)} - \frac{\eta_t}{r}\left(G_{n,t}^{(e)}a_{k,t}^{(e)}(a_{k,t}^{(e)})^T + b_{k,t}^{(e)}(b_{k,t}^{(e)})^T G_{n,t}^{(e)}\right)\\
        &+ \frac{\eta_t^2}{r^2}\left(G_{n,t}^{(e)}a_{k,t}^{(e)} (b_{k,t}^{(e)})^T G_{n,t}^{(e)}\right)\nonumber\\
        \Delta w_{k,t}^{(e)} &\coloneq - \frac{\eta_t}{r}\left(G_{n,t}^{(e)}a_{k,t}^{(e)}(a_{k,t}^{(e)})^T + b_{k,t}^{(e)}(b_{k,t}^{(e)})^T G_{n,t}^{(e)}\right)\label{eqn:delta_w}\\
        &+ \frac{\eta_t^2}{r^2}\left(G_{n,t}^{(e)}a_{k,t}^{(e)} (b_{k,t}^{(e)})^T G_{n,t}^{(e)}\right)\nonumber
    \end{align}

    Then the global update becomes

    \begin{align}
        W_{t+1} - W_{t} &= \frac{1}{L}\sum_{k=1}^{L}(w_{k,t}^{(E)} - w_{k,t}^{(0)})\\
        &= \frac{1}{L}\sum_{k=1}^L\sum_{e=0}^{E-1}\Delta w_{k,t}^{(e)}\label{eqn:delta_sum}\\
         W_{t+1} - W_{t} &= -\frac{\eta_t}{Lr}\sum_{k=1}^L\sum_{e=0}^{E-1}\left(G_{n,t}^{(e)}a_{k,t}^{(e)}(a_{k,t}^{(e)})^T + b_{k,t}^{(e)}(b_{k,t}^{(e)})^T G_{n,t}^{(e)}\right)\label{eqn:full_update}\\
        &+ \frac{\eta_t^2}{Lr^2}\sum_{k=1}^L\sum_{e=0}^{E-1}\left(G_{n,t}^{(e)}a_{k,t}^{(e)} (b_{k,t}^{(e)})^T G_{n,t}^{(e)}\right)\nonumber
    \end{align}

    To simplify the $G_{n,t}^{(e)}$ in the above, we take the following decomposition

    \begin{align}
        G_{n,t}^{(e)} &= \nabla f_n(W_{n,t}^{(e)})\\
        &= \nabla f(W_{t}) + \underbrace{(\nabla f_n(W_{t}) - \nabla f(W_{t}))}_{\zeta_{n,t}} + \underbrace{(\nabla f_n(W_{n,t}^{(e)}) - \nabla f_n(W_t))}_{\psi_{n,t}^{(e)}}\label{eqn:f_decomp}
    \end{align}

    Substituting Equation \ref{eqn:f_decomp} into \ref{eqn:full_update}

    \begin{align}
        W_{t+1} - W_{t} &= \underbrace{-\frac{\eta_t}{Lr} \sum_{k=1}^L\sum_{e=0}^{E-1} \left(\nabla f(W_t) a_{k,t}^{(e)} (a_{k,t}^{(e)})^T + b_{k,t}^{(e)} (b_{k,t}^{(e)})^T \nabla f(W_t) \right)}_{\mathcal{A}_{t}}\\
        &+  \underbrace{-\frac{\eta_t}{Lr} \sum_{k=1}^L\sum_{e=0}^{E-1} \left(\zeta_{n,t} a_{k,t}^{(e)} (a_{k,t}^{(e)})^T + b_{k,t}^{(e)} (b_{k,t}^{(e)})^T \zeta_{n,t} \right)}_{\mathcal{B}_{t}}\nonumber\\
        &+ \underbrace{-\frac{\eta_t}{Lr} \sum_{k=1}^L\sum_{e=0}^{E-1} \left(\psi^{(e)}_{n,t} a_{k,t}^{(e)} (a_{k,t}^{(e)})^T + b_{k,t}^{(e)} (b_{k,t}^{(e)})^T \psi^{(e)}_{n,t} \right)}_{\mathcal{C}_{t}}\nonumber\\
        &+ \underbrace{\frac{\eta_t^2}{Lr^2}\sum_{k=1}^L\sum_{e=0}^{E-1}\left(G_{n,t}^{(e)}a_{k,t}^{(e)} (b_{k,t}^{(e)})^T G_{n,t}^{(e)}\right)}_{\mathcal{D}_t}\nonumber\label{eqn:decomposed_global}
    \end{align}

    From $L_{f}$-smoothness and applying Equation \ref{eqn:decomposed_global} and then taking an expectation 

    \begin{align}
        f(W_{t+1}) &\leq f(W_{t}) + \langle \nabla f(W_t), W_{t+1} - W_t\rangle + \frac{L_f}{2}\|W_{t+1} - W_t\|_F^2\\
        f(W_{t+1}) - f(W_t) &\leq \mathrm{Tr}(\nabla f(W_t)^T(W_{t+1} - W_{t})) + \frac{L_f}{2}\|W_{t+1} - W_t\|_F^2\\
        &\leq \mathrm{Tr}(\nabla f(W_t)^T\mathcal{A}_t) + \mathrm{Tr}(\nabla f(W_t)^T\mathcal{B}_t) + \mathrm{Tr}(\nabla f(W_t)^T\mathcal{C}_t) \\
        &+ \mathrm{Tr}(\nabla f(W_t)^T\mathcal{D}_t) +\frac{L_f}{2}\|W_{t+1} - W_t\|_F^2\nonumber\\
        \mathbb{E}[f(W_{t+1}) - f(W_t)]&\leq \mathrm{Tr}(\nabla f(W_t)^T\mathbb{E}[\mathcal{A}_t]) + \mathrm{Tr}(\nabla f(W_t)^T\mathbb{E}[\mathcal{B}_t]) + \mathrm{Tr}(\nabla f(W_t)^T\mathbb{E}[\mathcal{C}_t])\label{eqn:grand_combination} \\
        &+ \mathrm{Tr}(\nabla f(W_t)^T\mathbb{E}[\mathcal{D}_t])+ \frac{L_f}{2}\mathbb{E}[\|W_{t+1} - W_t\|_F^2]\nonumber
    \end{align}

Simplifying each of the $\mathcal{A}_t, \mathcal{B}_t, \mathcal{C}_t$ terms individually 

\begin{align}
    \mathrm{Tr}(\nabla f(W_t)^T\mathcal{A}_t) &= -\frac{\eta_t}{Lr} \sum_{k=1}^L\sum_{e=0}^{E-1} \mathrm{Tr}(\nabla f(W_t)^T\nabla f(W_t) a_{k,t}^{(e)} (a_{k,t}^{(e)})^T) \\
    &- \frac{\eta_t}{Lr} \sum_{k=1}^L\sum_{e=0}^{E-1} \mathrm{Tr}(\nabla f(W_t)^Tb_{k,t}^{(e)} (b_{k,t}^{(e)})^T \nabla f(W_t))\nonumber\\
    &= -\frac{\eta_t}{Lr} \sum_{k=1}^L\sum_{e=0}^{E-1}\langle \nabla f(W_t), \nabla f(W_t) a_{k,t}^{(e)} (a_{k,t}^{(e)})^T) + b_{k,t}^{(e)} (b_{k,t}^{(e)})^T\nabla f(W_t) \rangle\\ 
    &= -\frac{\eta_t}{Lr} \sum_{k=1}^L\sum_{e=0}^{E-1}\langle \nabla f(W_t), \mathcal{M}_{k,t}^{(e)}(\nabla f(W_t))\rangle
\end{align}

Denoting $\Delta \mathcal{M}_{k,t}^{(e)}=  \mathcal{M}_{k,t}^{(e)} -  \mathcal{M}_{k,t}^{(0)}$ and using the previously obtained bound Equation \ref{eqn:operator_drift} on the operator drift along with the Persistent Isometry Lemma \ref{lemma:isometry}

\begin{align}
    \mathrm{Tr}(\nabla f(W_t)^T\mathcal{A}_t) &= -\frac{\eta_t}{Lr} \sum_{k=1}^L\sum_{e=0}^{E-1}(\langle \nabla f(W_t), \mathcal{M}_{k,t}^{(0)}(\nabla f(W_t))\rangle\\
    &- (\langle \nabla f(W_t), \Delta\mathcal{M}_{k,t}^{(e)}(\nabla f(W_t))\rangle)\nonumber\\
    &\leq -\frac{\eta_t}{Lr} \sum_{k=1}^L\sum_{e=0}^{E-1} \gamma \|\nabla f(W_t)\|_F^2 + \frac{\eta_t}{Lr} \sum_{k=1}^L\sum_{e=0}^{E-1}\frac{4\mathcal{C}\eta_t EG}{r}\|\nabla f(W_t)\|_F^2
\end{align}

where the last term changes sign due to considering worst case of the spectral norm given in Equation \ref{eqn:operator_drift}. 

\begin{align}
    \mathrm{Tr}(\nabla f(W_t)^T\mathbb{E}[\mathcal{A}_t])\leq (-\frac{\gamma\eta_t}{r}+\frac{4\mathcal{C}\eta_t^2E^2G}{r}) \|\nabla f(W_t)\|_F^2\label{eqn:A_bound}
\end{align}

Taking permutation symmetry, and noting that 

\begin{align}
    \mathbb{E}_{\pi(t)}[\zeta_{n,t}] &= \mathbb{E}_n[\nabla f_n(W_t) - \nabla f(W_t)]\\
    &= (\frac{1}{N}\sum_{n=1}^N \nabla f_n(W_t) )- \nabla f(W_t)\\
     \mathbb{E}_{\pi(t)}[\zeta_{n,t}] &= 0 \text{ , by definition of $f(W_t)$}
\end{align}

Using Equation \ref{eqn:e_residual_bound}, we can derive analogous bound to Equation \ref{eqn:residual_bound} but for $e\leq E$,

\begin{align}
    \|b_{k,t}^{(e)}(b_{k,t}^{(e)})^T-b_{k,t}(b_{k,t})^T\|_2\leq \frac{2\mathcal{C}\eta_teG}{r}\label{eqn:e_full_residual_bound}
\end{align}

Define the residual $R^{b}_{e,k,t}\coloneq b_{k,t}^{(e)}(b_{k,t}^{(e)})^T-b_{k,t}(b_{k,t})^T$ so that Equation \ref{eqn:e_full_residual_bound} shows $\|R^{b}_{e,k,t}\|\leq \frac{2\mathcal{C}\eta_teG}{r}$. Define $R_{e,k,t}^{a}$ analogously. Since on round $t$, $b_{k,t},a_{k,t}$ are determined and independent of the permutation, we can perform the following separation

\begin{align}
     \mathrm{Tr}(\nabla f(W_t)^T\mathbb{E}[\mathcal{B}_t]) &= -\mathrm{Tr}\Big[\nabla f(W_t)^T\frac{\eta_t}{Lr} \sum_{k=1}^L\sum_{e=0}^{E-1} \left(\mathbb{E}[\zeta_{n,t} a_{k,t}^{(e)} (a_{k,t}^{(e)})^T] + \mathbb{E}[b_{k,t}^{(e)} (b_{k,t}^{(e)})^T \zeta_{n,t}] \right)\Big]\\
     &= -\mathrm{Tr}\Big[\nabla f(W_t)^T\frac{\eta_t}{Lr} \sum_{k=1}^L\sum_{e=0}^{E-1}\Big(\mathbb{E}[\zeta_{n,t}] (a_{k,t} (a_{k,t})^T +  b_{k,t} (b_{k,t})^T\mathbb{E}[\zeta_{n,t}]\Big)\Big]
     \\&- \mathrm{Tr}\Big[\nabla f(W_t)^T\frac{\eta_t}{Lr}\sum_{k=1}^L\sum_{e=0}^{E-1}\mathbb{E}[\zeta_{n,t}R_{e,k,t}^a + R_{e,k,t}^b \zeta_{n,t}]\Big]\nonumber
\end{align}


The first term vanishes due to the permutation symmetry and $\mathbb{E}[\zeta_{n,t}]=0$. To bound the second term, we use the identities $|\mathrm{Tr}(X^TY)|\leq \|X\|_F\|Y\|_F$, $\|ZR\|_F\le\|Z\|_F\|R\|_2$, and $\|RZ\|_F\le\|R\|_2\|Z\|_F$, 

\begin{align}
    \big|\mathrm{Tr}\big(\nabla f(W_t)^T(\zeta_{n,t} R_{e,k,t}^a+R^b_{e,k,t}\zeta_{n,t})\big)\big|&\leq\|\nabla f(W_t)\|_F\,\|\zeta_{n,t}\|_F\,\big(\|R_{e,k,t}^a\|_2+\|R_{e,k,t}^b\|_2\big)
\end{align}

From Triangle Inequality and Assumption B.2 we can show $\|\nabla f(W_t)\|_F\leq G$ and $\|\zeta_{n,t}\|\leq \|\nabla f_n(W_t)\|_F + \|\nabla f(W_t)\|_F\leq 2G$. Using Equation \ref{eqn:e_full_residual_bound}, we can derive that 

\begin{align}
       \big|\mathrm{Tr}\big(\nabla f(W_t)^T(\zeta_{n,t} R_{e,k,t}^a+R^b_{e,k,t}\zeta_{n,t})\big)\big|&\leq G \cdot 2G\cdot \frac{4\mathcal{C}e\eta_tG}{r}=\frac{8\mathcal{C}e\eta_tG^3}{r}\\
       |\mathrm{Tr}(\nabla f(W_t)^T\mathbb{E}[\mathcal{B}_t])| &\leq \Big|\frac{\eta_t}{Lr}\sum_{k=1}^L\sum_{e=0}^{E-1}\frac{8\mathcal{C}e\eta_tG^3}{r} \Big|\\
       &\leq \frac{8\mathcal{C}\eta_t^2G^3}{r^2}\sum_{e=0}^{E-1}e\\
       |\mathrm{Tr}(\nabla f(W_t)^T\mathbb{E}[\mathcal{B}_t])|&\leq \frac{4E(E-1)\mathcal{C}\eta_t^2G^3}{r^2}\label{eqn:B_bound}
\end{align}

In other words, permutation symmetry mitigates the heterogeneity term up to a second order error, which vanishes when $E=1$. 

To simplify the $\mathcal{C}_t$ and the quadratic Lipschitz terms, we first consider a bound on $\Delta w_{k,t}^{(e)}$ in Equation \ref{eqn:delta_w}. 

\begin{align}
    \|\Delta w_{k,t}^{(e)}\|_2 &\leq \frac{2\eta_t \mathcal{C}G}{r} + \frac{\eta_t^2\mathcal{C}G^2}{r^2}\\
    &\leq \frac{\eta_t\mathcal{C}G}{r}(2 + \eta_t\frac{G}{r})
\end{align}

Choosing learning rate $\eta_t\leq \frac{r}{G}$ gives

\begin{align}
    \|\Delta w_{k,t}^{(e)}\|_2 \leq \frac{3\eta_t \mathcal{C}G}{r}\label{eqn:delta_bound}
\end{align}

Using the identity $\mathrm{rank}(AB)\leq\min(\mathrm{rank}(A), \mathrm{rank}(B))$ and $\mathrm{rank}(A+B)\leq \mathrm{rank}(A)+ \mathrm{rank}(B)$, we can identify that $\Delta w_{k,t}^{(e)}$ in Equation \ref{eqn:delta_w} is at most rank $3$: The first two terms are products of $G_{n,t}^{(e)}$ and rank-1 outer products, while the third term can be rewritten as an outer product $(G_{n,t}^{(e)}a_{k,t}^{(e)})((G_{n,t}^{(e)})^Tb_{k,t}^{(e)})^T$. Then we can convert the spectral norm in Equation \ref{eqn:delta_bound} into a Frobenius norm,

\begin{align}
    \|\Delta w_{k,t}^{(e)}\|_F\leq \sqrt{\mathrm{rank}(\Delta w_{k,t}^{(e)})} \|\Delta w_{k,t}^{(e)}\|_2\leq \frac{3\sqrt{3}\eta_t \mathcal{C}G}{r}
\end{align}

Using Triangle Inequality and combining Equations \ref{eqn:delta_bound} and \ref{eqn:delta_sum}

\begin{align}
    \|W_{t+1} - W_{t}\|_F&\leq \frac{1}{L}\sum_{k=1}^L\sum_{e=0}^{E-1}\|\Delta w_{k,t}^{(e)}\|_F\ \\
    &\leq \frac{3\sqrt{3}\eta_t\mathcal{C} EG}{r}\\
    \frac{L_f}{2}\|W_{t+1} - W_{t}\|_F^2&\leq \frac{27L_f\eta_t^2\mathcal{C}^2E^2G^2}{2r^2}\label{eqn:lipscthiz}
\end{align}

Using $L_{f}$-smoothness of $f_n$, 

\begin{align}
    \|\psi^{(e)}_{n,t}\|_F &= \|\nabla f_n(W_{n,t}^{(e)}) - \nabla f_n(W_t)\|_F\\
    &\leq L_f\|W_{n,t}^{(e)} - W_t\|_F\\
    &\leq L_f\|\frac{1}{r}\sum_{k\in S_{n,t}} b_{k,t}^{(e)} (a_{k,t}^{(e)})^T - \frac{1}{L}\sum_{k=1}^L b_{k,t} (a_{k,t})^T\|_F\\
    &\leq L_f\|W_{n,t}^{(e)} - W_{n,t}^{(0)}\|_F+ L_f\|W_{n,t}^{(0)} - W_t\|_F\\
    &\leq L_f \sum_{i=0}^{e-1}\frac{1}{r}\sum_{k\in S_{n,t}}\|\Delta w_{k,t}^{(i)}\| + L_f\|W_{n,t}^{(0)} - W_t\|_F\\
    &\leq \frac{3\sqrt{3}L_f e\eta_t \mathcal{C}G}{r} + L_f\|W_{n,t}^{(0)} - W_t\|_F\\
    \|\psi^{(e)}_{n,t}\|_F&\leq \frac{3\sqrt{3}L_f E\eta_t \mathcal{C}G}{r} + L_f\|W_{n,t}^{(0)} - W_t\|_F
\end{align}

Let us focus on the second term of this inequality. This represents a "local blindness" error $\xi_{n,t}$ arising because the local rank-$r$ basis set deviates from the global rank $L$ library. 

By our Bounded basis norms Assumption \ref{ass:basis_norm} gives $\|w_{k,t}\|_F\leq \|w_{k,t}\|_2=\|b_{k,t}(a_{k,t})^T\|_2\leq \mathcal{C}$, since $w_{k,t}$ is rank $1$. 

\begin{align}
    \xi_{n,t} &\coloneq \|W_{n,t}^{(0)} - W_t\|_F \\
    &= \|\frac{1}{r}\sum_{k\in S_{n,t}} w_{k,t} - \frac{1}{L}\sum_{k=1}^L w_{k,t}\|_F\\
    &= \|(\frac{1}{r} - \frac{1}{L})\sum_{k\in S_{n,t}}w_{k,t} - \frac{1}{L}\sum_{k\notin S_{n,t}}w_{k,t}\|_F\\
    &\leq (\frac{1}{r}-\frac{1}{L})r\mathcal{C} + \frac{1}{L}(L-r)\mathcal{C}\\
    \xi_{n,t}  &\leq 2\mathcal{C}(1-\frac{r}{L})
\end{align}

\begin{align}
    \|\psi^{(e)}_{n,t}\|_F &\leq \frac{3\sqrt{3}L_f E\eta_t \mathcal{C}G}{r} + 2L_f\mathcal{C}(1-\frac{r}{L})
\end{align}

\begin{align}
    \mathrm{Tr}(\nabla f(W_t)^T\mathcal{C}_t) &= -\frac{\eta_t}{Lr} \sum_{k=1}^L\sum_{e=0}^{E-1} \mathrm{Tr}\left[\nabla f(W_t)^T\left(\psi^{(e)}_{n,t} a_{k,t}^{(e)} (a_{k,t}^{(e)})^T + b_{k,t}^{(e)} (b_{k,t}^{(e)})^T \psi^{(e)}_{n,t} \right)\right]\\
    &\leq \frac{\eta_t}{Lr}\sum_{k=1}^L\sum_{e=0}^{E-1} \|\nabla f(W_t)\|_F (2\mathcal{C})\|\psi_{n,t}^{(e)}\|_F\\
    &\leq \frac{\eta_t}{Lr}\sum_{k=1}^L\sum_{e=0}^{E-1} \left[\frac{6\sqrt{3}L_f E\eta_t \mathcal{C}^2G^2}{r} + 4L_f\mathcal{C}^2G(1-\frac{r}{L})\right]\\
    \mathrm{Tr}(\nabla f(W_t)^T\mathcal{C}_t) &\leq \frac{6\sqrt{3}L_f E^2\eta_t^2 \mathcal{C}^2G^2}{r^2} + 4L_fE\eta_t\mathcal{C}^2G(\frac{1}{r}-\frac{1}{L})\label{eqn:C_bound}
\end{align}

Finally to bound the $\mathcal{D}_t$ term, 

\begin{align}
    \mathrm{Tr}(\nabla f(W_t)^T\mathcal{D}_t) &= \frac{\eta_t^2}{Lr^2}\sum_{k=1}^L\sum_{e=0}^{E-1} \mathrm{Tr}(\nabla f(W_t)^T G_{n,t}^{(e)}a_{k,t}^{(e)} (b_{k,t}^{(e)})^T G_{n,t}^{(e)})\\
    &\leq \frac{\eta_t^2}{Lr^2}\sum_{k=1}^L\sum_{e=0}^{E-1} G^3\mathcal{C}\\
    \mathrm{Tr}(\nabla f(W_t)^T\mathcal{D}_t) &\leq \frac{\eta_t^2 G^3 E\mathcal{C}}{r^2}\label{eqn:D_bound}
\end{align}

Combining Equations \ref{eqn:A_bound}, \ref{eqn:B_bound}, \ref{eqn:C_bound}, \ref{eqn:D_bound}, and \ref{eqn:lipscthiz} into \ref{eqn:grand_combination}

\begin{align}
    \mathbb{E}[f(W_{t+1}) - f(W_t)]&\leq (-\frac{\gamma\eta_t}{r}+\frac{4\mathcal{C}\eta_t^2E^2G}{r}) \|\nabla f(W_t)\|_F^2\\
    &+ \frac{4E(E-1)\mathcal{C}\eta_t^2G^3}{r^2}\nonumber\\
    &+ \frac{6\sqrt{3}L_f E^2\eta_t^2 \mathcal{C}^2G^2}{r^2} + 4L_fE\eta_t\mathcal{C}^2G(\frac{1}{r}-\frac{1}{L})\nonumber\\
    &+ \frac{\eta_t^2 G^3 E\mathcal{C}}{r^2}\nonumber\\
    &+ \frac{27L_f\eta_t^2\mathcal{C}^2E^2G^2}{2r^2} \nonumber\\
     \mathbb{E}[f(W_{t+1}) - f(W_t)] &\leq  (-\frac{\gamma\eta_t}{r}+\frac{4\mathcal{C}\eta_t^2E^2G}{r}) \|\nabla f(W_t)\|_F^2 + 4L_fE\eta_t\mathcal{C}^2G(\frac{N-1}{Nr})\label{eqn:descent_lemma} \\
     &+ \frac{\eta_t^2\big(12\sqrt3L_fE^2\mathcal{C}^2G^2+2G^3E\mathcal{C}+27L_fG^2E^2\mathcal{C}^2+8E(E-1)\mathcal{C}G^3\big)}{2r^2}\nonumber
\end{align}

From the definition, $\eta_t\leq \frac{\gamma}{8\mathcal{C}GE^2}$ ensures the coefficient, specifically $-\frac{\gamma\eta_t}{2r}$ of $\|\nabla f(W_t)\|_F^2$ remains negative for the descent lemma shown in Equation \ref{eqn:descent_lemma}. 

let $B\coloneq \frac{4L_fE\mathcal{C}^2G(N-1)}{Nr}$. After $T$ rounds, with $f^* \leq f(W_{T})$, we have 

\begin{align}
    \mathbb E[f(W_{t+1})]-\mathbb E[f(W_t)]&\leq-\frac{\gamma\eta_t}{2r}\,\mathbb E\|\nabla f(W_t)\|_F^2+B\,\eta_t+\frac{K}{2r^2}\,\eta_t^2 \\
    \mathbb E[f(W_T)]-f(W_0)&\leq-\frac{\gamma}{2r}\sum_{t=0}^{T-1}\eta_t\,\mathbb E\|\nabla f(W_t)\|_F^2+B\sum_{t=0}^{T-1}\eta_t+\frac{K}{2r^2}\sum_{t=0}^{T-1}\eta_t^2
\end{align}

Using $f^*\leq f(W_T)$,

\begin{align}
    \frac{\gamma}{2r}\sum_{t=0}^{T-1}\eta_t\,\mathbb E\|\nabla f(W_t)\|_F^2\le f(W_0)-f^*+B\sum_{t=0}^{T-1}\eta_t+\frac{K}{2r^2}\sum_{t=0}^{T-1}\eta_t^2
\end{align}

Since $\eta_t=\frac{\eta_0}{\sqrt{t+1}}\ge\frac{\eta_0}{\sqrt T}$,

\begin{align}
    \frac{\gamma\eta_0}{2r\sqrt T}\sum_{t=0}^{T-1}\mathbb E\|\nabla f(W_t)\|_F^2\le f(W_0)-f^*+B\sum_{t=0}^{T-1}\eta_t+\frac{K}{2r^2}\sum_{t=0}^{T-1}\eta_t^2
\end{align}

Applying the following identities  

\begin{align}
    \sum_{t=0}^{T-1}\eta_t&=\eta_0\sum_{s=1}^{T}\frac1{\sqrt s}\le2\eta_0\sqrt T\\
    \sum_{t=0}^{T-1}\eta_t^2&=\eta_0^2\sum_{s=1}^{T}\frac1s\le\eta_0^2(1+\ln T)
\end{align}

And multiplying both sides by $\frac{2r}{\gamma\eta_0\sqrt T}$

\begin{align}
    \frac1T\sum_{t=0}^{T-1}\mathbb E\|\nabla f(W_t)\|_F^2&\le\frac{2r(f(W_0)-f^*)}{\gamma\eta_0\sqrt T}+\frac{2r}{\gamma\eta_0\sqrt T}\cdot B\cdot2\eta_0\sqrt T\\
    &+\frac{2r}{\gamma\eta_0\sqrt T}\cdot\frac{K}{2r^2}\cdot\eta_0^2(1+\ln T)\nonumber
\end{align}

Then substituting back $B$, we get the desired result

\begin{align}
    \frac1T\sum_{t=0}^{T-1}\mathbb E\|\nabla f(W_t)\|_F^2\le\frac{2r(f(W_0)-f^*)}{\gamma\eta_0\sqrt T}+\frac{16L_fE\mathcal{C}^2G(N-1)}{N\gamma}+\frac{K\eta_0(1+\ln T)}{\gamma r\sqrt T}
\end{align}

Rewriting the above in Big-O notation

\begin{align}
    \frac1T\sum_{t=0}^{T-1}\mathbb E\|\nabla f(W_t)\|_F^2\le\frac{2r(f(W_0)-f^*)}{\gamma\eta_0\sqrt T}+\frac{16L_fEC^2G(N-1)}{N\gamma}+O\Big(\frac{\eta_0\log T}{\gamma r\sqrt T}\Big)
\end{align}

\end{proof}

\subsection{Analysis and limitations}\label{sec:theory_limit}

We note a few differences of this proof from the main algorithm shown in Algorithm \ref{alg-LEAN}. We have simplified the setting to a synchronous full participation case and did not include the proximal term in the analysis. However, we stress that even in this limited scope we have demonstrated important mechanisms in LEAN that distinguish it from other FL algorithms. 

Notably we do not require a separate bounded-dissimilarity assumption common in some previous FL works \cite{FLoRA, fedavg_converge, ist}. Our \emph{shuffling} mechanism cancels the heterogeneity term exactly at the start of each round, leaving only an $O(\eta_t^2)$ residual from within-round drift (Equation \ref{eqn:B_bound}), which vanishes when $E=1$.” While the convergence result in Theorem \ref{thm:converge} has a minimal noise floor that is independent of $\eta_t$, it is notably not directly dependent on data heterogeneity. We also remark that this noise floor is very pessimistic; our "local blindness" error that generated this term used worst case bounds. For example, in the case of $E=1$ where there is no local drift, we can leverage permutation symmetry to also simplify the $\psi_{n,t}^{(e)}$ term since the single round of gradient descent preserves its linearity, but this is not reflected in the convergence theorem. 

The library size bound in Lemma \ref{lemma:isometry} greatly exceeds the library sizes used in experiments, e.g. for $a=b=1280$ (ViT-H dimensions) and $\delta=0.01$, the required $L$ is approximately $1.9\times 10^5$, whereas our experimental library sizes range from 1,280 to 12,800. However this is only a sufficient condition for the lemma and as seen in our experiments is often unnecessary in practice. 
On the other hand, consider the kernel space of the initial operator, i.e. $g\in \mathbb{R}^{b\times a}$ such that  $\mathcal{M}_{k,0}(g)=0$. Since  $\mathcal{M}_{k,0}$ is positive semi-definite, this is only satisfied when $ga=0$ and $b^Tg=0$, i.e. every row of $g$ is orthogonal to $a$ and every column orthogonal to $b$. Hence the kernel space is of dimension $(b-1)(a-1)$, for $b_{k,0},a_{k,0}\neq 0$ which holds almost surely, so $rank(\mathcal{M}_{k,0})=ba - (b-1)(a-1)=a+b-1$. Since rank is sub-additive,

\begin{align}
    rank(\mathcal{M}_0) &= rank(\sum_{k=1}^L\frac{1}{L}\mathcal{M}_{k,0})\\
    &\leq \sum_{k=1}^L rank(\mathcal{M}_{k,0})\\
    rank(\mathcal{M}_0) &\leq L(b+a-1)
\end{align}

$\lambda_{min}(\mathcal{M}_0)>0$ is a necessary condition of Lemma \ref{lemma:isometry} as $\mathcal{M}_0$ needs to be full-rank $ba$; otherwise, for some $g\neq 0$ $\mathcal{M}_0(g)=0$ and no $\gamma>0$ can satisfy $\langle g, \mathcal{M}_t(g)\rangle\geq \gamma\|g\|_F^2$. Therefore we can derive a necessary bound on $L$

\begin{align}
    rank(\mathcal{M}_0)&=ba\\
    L&\geq \frac{ba}{b+a-1}\\
    L &> \frac{1}{1/b+1/a}\label{eqn:necessary_lbound}
\end{align}

On ViT-H, this bound requires $L>640$, which all our experiments satisfy, as the smallest library uses 1,280 bases. 


\section{Additional Baselines and Experiments}

\subsection{SLoRA}

\textit{SLoRA} \citep{slora} performs an initial warm-up stage with a sample of clients via sparsified full fine-tuning \citep{sparse_ft} and then the initialization for LoRA adapters is determined by a rank-$r$ SVD approximation of the full-rank update. In \citep{slora}, SLoRA is implemented with federated fine-tuning by FedAvg; however, we note that SLoRA's sparse warm-up is compatible with LEAN as a first initialization of the basis library.  

\begin{table}[ht]
\centering
\caption{SLoRA Comparison with LEAN and FLoRA}
\begin{tabular}{ccc}
Method & Acc. (\%) & Bytes (GB)\\
\hline
FLoRA & 71.30 & $(261.6)^{BT}$ \\
LEAN & 94.26 & 100.72 \\
FedAvg w/ SLoRA & 98.40 & 298.78 \\
\end{tabular}
\label{tab:slora}
\end{table}

We present the results in Table \ref{tab:slora} on the Flowers dataset with 25 participants, high asynchrony, and light data heterogeneity. FedAvg w/ SLoRA initialization makes significant accuracy improvements over base FedAvg (see Table \ref{tab:baseline_compare}) as a result of the better initialization, exceeding LEAN in final accuracy but requiring approximately $3\times$ the communication. Note that LEAN's bytes differ from Table \ref{tab:25cli_heavyasync} because the $95\%$ threshold is set by FedAvg w/ SLoRA here. We recognize that SLoRA initialization can significantly improve model performance; however, we again emphasize that it is fully orthogonal to our work.

\subsection{Alternative checkOut Sampling}

In Algorithm \ref{alg-LEAN} line $15$, we do not provide a specific protocol for sampling the bases during \textsc{checkOut}. Our main experiments used uniformly random sampling in implementation. However, we present two alternative approaches for sampling bases:

\textbf{Staleness-Aware} - We compute a timestamp difference between the current time of \textsc{checkOut} and the time of last \textsc{checkIn} for an available parameter pair in the library. We compute a softmax probability distribution and prioritize less stale parameters when sampling.  

\textbf{Similarity-Aware} - At each basis pair we record the class label distribution of the site checking-in their basis pairs. During checkout we compute the cosine similarity between the current site checking-out and the recorded distributions of the pairs in the library. Then we compute a softmax probability distribution from the cosine similarities, prioritizing the most similar distributions. We note that similarity-aware \textsc{checkOut} has severe problems with privacy preservation as it relies on sharing the local data distributions to the central server. 

\begin{table}[t]
\centering
\caption{Sampling Method Comparison}
\begin{tabular}{ccc}
\hline
Method & Acc. (\%) & Bytes (GB)\\
\hline
Random & 94.26 & 88.09 \\
Similarity-aware & 94.13 & 90.85 \\
Staleness-aware & 92.81 & 90.92
\end{tabular}
\label{tab:routing}
\end{table}

Table \ref{tab:routing} compares these two sampling methods to uniform randomness for the high asynchrony, light heterogeneity case on 25 participants and Flowers dataset. Similarity-aware sampling has similar accuracies as uniformly random routing while staleness-aware performs slightly worse. A deeper exploration of the impact of sampling protocol is left for future work, but these results justify our choice of using the simplest, uniform random sampling for basis selection.   

\section{Detailed Limitations}\label{sec:detail_limits}

\subsection{Convergence Proof}

As discussed in Section \ref{sec:theory_limit}, the convergence analysis diverges from the experiments as the former is limited to the synchronous setting without a proximal term. In particular, the bound from Theorem \ref{thm:converge} holds for a fixed horizon with a horizon-dependent step size, whereas a weak federation trains indefinitely and may deploy the model at any time. We emphasize that anytime behavior in this setting is empirical. The analysis primarily serves as a sanity check that LEAN's operations are fundamentally sound and not as a prescription for practical implementation, where some assumptions may not hold. An exception is Equation \ref{eqn:necessary_lbound} which sets a necessary, model-dependent minimum library size but we note that this bound is relatively small and easily satisfied in practice. 

A more careful theoretical analysis in the asynchronous setting is highly nontrivial. Prior works in asynchronous federated learning \citep{fedfa, fedasync, fedbuff} assume a bounded staleness or delay in their analyses, which does not hold in cases like our Pareto-distributed staleness. Hence, we leave the asynchronous convergence analysis to future work. 

\subsection{Additional Asynchronous Settings}

While inherent asynchrony is highlighted as a core advantage of LEAN, this paper does not directly address asynchronous issues such as stragglers and instead mitigates them with a large library size. Extreme long-tailed and adversarial site schedules were not tested. In particular, the flexible library size permits adversarial site behaviors that are not normally possible in other federated settings, such as sudden increases in library size followed by underutilization, which can diminish model performance. Such behaviors can also naturally arise in cases of high variance site delays. The scope of this work is primarily in proposing the LEAN algorithm and empirically testing it in the \emph{weak federation} setting, which includes asynchrony but not necessarily in the most pathological cases. 

\subsection{Threat Model}

We recognize that the threat model in our system is non-adversarial as we consider only honest client updates of weak federations that are sent to a trustworthy server. Byzantine clients \citep{byzantine_bucketing, byzantine_robust, byzantine_experimental, krum} are outside this scope, particularly since robust aggregation methods trade off against heterogeneous data, our main focus.

The basis shuffling mechanism may provide some degree of privacy by decoupling and anonymizing the full LoRA $B,A$ updates into individual rank-1 pairs. However, the rank-1 pairs still represent data leakage to both other clients and to the server. Differential privacy \citep{federated_dp} may mitigate this privacy leakage, but more careful empirical testing is needed.

\section{Additional Ablations}

\subsection{Single-Run Stability and Selection-Free Reporting}\label{sec:model_selection}

Because a model in a weak federation may be deployed at any time without checkpoint selection, as discussed in Section \ref{sec:floraprox}, the relevant quantity is the distribution of accuracy over training rather than its peak. Our nine-model window criterion in Table \ref{tab:floraprox} is itself a mild form of selection, since it picks the best window. Moreover, because in practical deployed training, the system cannot be simply restarted, we must consider the stability of this accuracy distribution within a single run. To address both these concerns, for a single run (seed 926) on the Birds dataset with 10 participants and high heterogeneity, Table \ref{tab:selection_free} reports the \emph{selection-free statistics}: the mean, standard deviation, and minimum over all check-ins after a fixed burn-in of $K\in \{0,50,100,200,400\}$ check-ins. 

As $K$ increases, LEAN has a monotonically increasing mean and minimum with decreasing standard deviation, indicating well-behaved model convergence behavior: Longer training results in better and more stable performance. By contrast, although FLoRAProx's mean is increasing with $K$, the consistently high standard deviation and low minimum until $K=400$ indicate FLoRAProx continues to produce low quality inference models throughout most of its training. These results reinforce our conclusion in Section \ref{sec:floraprox} that FLoRAProx struggles with stable and consistent performance. 

\begin{table}[h]
    \centering
    \footnotesize
    \vspace{-10 pt}
    \caption{Selection-Free LEAN vs. FLoRAProx, seed 926}
    \begin{tabular}{ccccccc}
    \hline
    K & \multicolumn{3}{c}{LEAN} & \multicolumn{3}{c}{FLoRAProx}\\
    \cline{1-7}
    & Mean & Std Dev. & Min & Mean & Std Dev. & Min \\
    0 & 18.05 & 6.21 & 1.00 & 7.74 & 9.11 & 0.25 \\
    50 & 18.74 & 5.30 & 2.25 & 8.37 & 9.23 & 0.25 \\
    100 & 19.46 & 4.41 & 9.25 & 9.14 & 9.33 & 0.25 \\
    200 & 20.48 & 3.47 & 10.10 & 11.22 & 9.28 & 0.25 \\
    400 & 21.65 & 2.71 & 15.01 & 17.93 & 7.11 & 4.25
    \end{tabular}
    \label{tab:selection_free}
\end{table}

\subsection{Proximal Term}

To study the effect of the proximal term, we test $\lambda$ sensitivity under high heterogeneity ($\rho=0.01$, Birds) and light heterogeneity ($\rho = 1.0$, Flowers). Because the two blocks use different datasets, their values are not directly comparable.

Increasing $\lambda$ can improve accuracy under
$\rho=0.01$ up to a point, but too large $\lambda$ results in a complete collapse. Under the $\rho=1.0$ regime, accuracy decreases as $\lambda$ grows up to a similar collapse point, which does suggest that there may be a heterogeneity-dependent optimum $\lambda$. We note, however, that in practice the heterogeneity of client data cannot be known a
priori, so our main experiments use a single fixed $\lambda$ across settings
with this practical limitation in mind. The sensitivity table lets
practitioners see the tradeoff.

\begin{table}[h]
    \centering
    \footnotesize
    \vspace{-10 pt}
    \caption{Proximal Term Sensitivity}
    \begin{tabular}{lcc}
    \hline
    $\lambda$ & Birds Acc.\ (\%), $\rho = 0.01$\\
    \midrule
    $3\times10^{-3}$ (default) & 25.39 \\
    $7\times 10^{-3}$ & 27.95 \\
    $1 \times 10^{-2}$ & 2.17  \\
    \hline
     & Flowers Acc.\ (\%), $\rho = 1.0$\\
     $3\times10^{-3}$ (default) &	95.18 \\
     $7\times10^{-3}$ &	94.23 \\
$1\times10^{-2}$ &	91.82 \\
     $3\times10^{-2}$ &	5.11 \\
    \end{tabular}
    \label{tab:proximal_term}
\end{table}

\section{Model and FL Settings}

\subsection{Federated Simulation}

All code was run on the PyTorch \citep{pytorch} (BSD-style) framework with arbitrary random seed $926$ on single seed runs and a trio of seeds $\{0, 111, 521\}$ for multiseed runs, except for LEAN on Birds with 10 participants and high heterogeneity in Table \ref{tab:floraprox}, which uses seeds $\{17, 111, 521\}$. 

In the federated simulations, \emph{LEAN} models were trained locally for $150$ batches before their weights are sent to the server while FLoRA, FedAvg, and FedProx were trained on $200$. An exception is for Qwen (Apache 2.0) models, where their local batches are $50$ for both LEAN and FLoRA. 

Specific algorithmic details in Section 6 for implementing our Dirichlet heterogeneity and asynchronous simulation are provided in this work's associated GitHub repository. 

FLOPs computation in simulations was aided by \emph{calflops} Python library (MIT License)(\url{https://pypi.org/project/calflops/0.0.2/}).  

All experiments were run on NVIDIA Tesla P100-SXM2-16GB GPUs on single nodes with 4 GPUs. We used a 16TB storage server to store model weights, datasets, and asynchronous execution plans.  

\subsection{Models}

The Vision Transformer implementation we used is based on \url{https://github.com/lukemelas/PyTorch-Pretrained-ViT} (no explicit license). 

Pre-trained model weights for H-14 ViT (\url{https://huggingface.co/google/vit-huge-patch14-224-in21k}) (Apache-2.0) and Qwen3-0.6B (\url{https://huggingface.co/Qwen/Qwen3-0.6B}) (Apache-2.0) were sourced from Huggingface. A script and configs are provided in our source code to adapt the H-14 weights to the ViT implementation we used.

ViT models were trained with a learning rate of $6 \times 10^{-3}$ and batch size of $8$. Qwen models were trained with a learning rate of $3\times 10^-4$ and batch size of $4$. 

In the case of proximal models \citep{fedprox}, we used a proximal coefficient of $3\times 10^{-3}$, which we found to be broadly effective. 

\subsection{Datasets}

The Oxford VGG 102-Flowers dataset \citep{flowers} was downloaded from the original website: \url{https://www.robots.ox.ac.uk/~vgg/data/flowers/102/} (no explicit license, used for non-commercial research)

CUB-Birds-200-2011 \citep{birds} was downloaded from the original website: \url{https://www.vision.caltech.edu/datasets/cub_200_2011/} (no explicit license, used for non-commercial research)

CIFAR100 \citep{cifar} (no explicit license, used for non-commercial research) was provided by the Python \emph{torchvision} library (BSD-3 Clause). 

We processed all image datasets with a $224\times 224$ center cropping to fit them to resolution of the ViT's pre-trained dataset.  

ChemLLMBench \citep{chembench} was downloaded from the original GitHub repository: \url{https://github.com/ChemFoundationModels/ChemLLMBench} (no explicit license, used for non-commercial research). We used the provided USPTO\_50k dataset \citep{uspto50k} (MIT license) for training and testing our models. Our prompt for the Qwen model was the following:

\begin{quote}
        You are an expert chemist. Given the reactants SMILES, your task is to predict the main product SMILES using your experienced chemical Reaction Prediction knowledge. Please strictly follow the format, no other information can be provided. You should only reply with SMILES string notations to represent the product. The input contains the reactants and reagents which are split by '.'. The product smiles must be valid and chemically reasonable. \\
        
        Reactants+Reagents: \\
        Products: 
\end{quote}



\end{document}